\documentclass{article}

\def\FIGDIR{./figures}          

\def\forsubmission{1} 

\usepackage{amsmath}
\usepackage{amsfonts}
\usepackage{amsthm}
\theoremstyle{definition}
\newtheorem{definition}{Definition}

\usepackage{newfloat}

\usepackage[small,compact]{titlesec}
\usepackage{algorithmic}
\usepackage[linesnumbered]{algorithm2e}
\usepackage{caption}
\usepackage{color}
\usepackage[normalem]{ulem}
\usepackage{microtype}
\usepackage{xspace}
\usepackage{graphicx}
\usepackage{booktabs} 

\usepackage{makecell}
\usepackage{multirow}
\usepackage{comment}
\usepackage{hhline}
\usepackage[para]{threeparttable} 
\makeatletter
\def\TPT@doparanotes{\par
   \prevdepth\z@ \TPT@hsize
   \TPTnoteSettings
   \parindent\z@ \pretolerance 8
   \linepenalty 200
   \renewcommand\item[1][]{\relax\ifhmode \begingroup
       \unskip
       \advance\hsize 10em 
       \penalty -45 \hskip\z@\@plus\hsize \penalty-19
       \hskip .07\hsize \penalty 9999 \hskip-.15\hsize
       \hskip .01\hsize\@plus-\hsize\@minus.01\hsize 
       \hskip 6em\@plus .3em
      \endgroup\fi
      \tnote{##1}\,\ignorespaces}%
   \let\TPToverlap\relax
   \def\endtablenotes{\par}%
}
\makeatother

\usepackage{hyperref}

\RequirePackage[nameinlink]{cleveref} 
\crefname{chapter}{Chapter}{Chapters}
\crefname{section}{Section}{Sections}
\crefname{subsection}{Subsection}{Subsections}
\crefname{equation}{Equation}{Equations}
\crefname{definition}{Definition}{Definitions}
\crefname{assumption}{Assumption}{Assumptions}
\crefname{theorem}{Theorem}{Theorems}
\crefname{figure}{Figure}{Figures}
\crefname{table}{Table}{Tables}
\crefname{BOX}{Box}{Boxes}
\let\autoref\cref 

\newcommand{\insertFigure}[2]{
    \begin{figure}[t]
        \centering
        \includegraphics[width=\linewidth]{\FIGDIR/#1.pdf}
        \vspace{-8mm}
        \caption{\small #2}
        \vspace{-4mm}
        \label{fig:#1}
    \end{figure}
}

\newcommand{\insertWideFigure}[2]{
    \begin{figure*}[h]
        \centering
        \includegraphics[width=\textwidth]{\FIGDIR/#1.pdf}
        \vspace{-8mm}
        \caption{\small #2}
        \vspace{-4mm}
        \label{fig:#1}
    \end{figure*}
}

\newcommand{\insertFormula}[3]{
\textbf{\small#1}
\\[-1.3\baselineskip]
\footnotesize
\revision{
\begin{align*}
#3
\end{align*}
}
\normalsize
\\[#2\baselineskip]
}

\DeclareFloatingEnvironment[
  fileext=lob,
  listname={List of Boxes},
  name=Box,
  placement=htp,
]{BOX}

\ifx\forsubmission\undefined
\newcommand{\TODO}[1]{\textcolor{red}{TODO: #1}}
\newcommand{\TK}[1]{\textcolor{blue}{TK: #1}}
\newcommand{\HK}[1]{\textcolor{cyan}{HK: #1}}
\newcommand{\LL}[1]{\textcolor{brown}{LL: #1}}
\newcommand{\CB}[1]{\textcolor{violet}{CB: #1}}
\newcommand{\MM}[1]{\textcolor{violet}{MM: #1}}
\newcommand{\VJ}[1]{\textcolor{blue}{VJ: #1}}
\newcommand{\JY}[1]{\textcolor{red}{JY: #1}}
\newcommand{\JS}[1]{\textcolor{green}{JS: #1}}
\newcommand{\revision}[1]{\textcolor{blue}{#1}}
\else
\newcommand{\TODO}[1]{\textcolor{red}{}}
\newcommand{\TK}[1]{\textcolor{blue}{}}
\newcommand{\HK}[1]{\textcolor{green}{}}
\newcommand{\LL}[1]{\textcolor{brown}{}}
\newcommand{\CB}[1]{\textcolor{violet}{}}
\newcommand{\MM}[1]{\textcolor{violet}{}}
\newcommand{\VJ}[1]{\textcolor{blue}{}}
\newcommand{\JY}[1]{\textcolor{red}{}}
\newcommand{\JS}[1]{\textcolor{green}{}}
\newcommand{\revision}[1]{#1}
\fi

\newcommand{\squishlist}{
 \begin{list}{$\bullet$}
  { \setlength{\itemsep}{0pt}
     \setlength{\parsep}{3pt}
     \setlength{\topsep}{3pt}
     \setlength{\partopsep}{0pt}
     \setlength{\leftmargin}{1.5em}
     \setlength{\labelwidth}{1em}
     \setlength{\labelsep}{0.5em} } }

\newcommand{\squishlisttwo}{
 \begin{list}{$\bullet$}
  { \setlength{\itemsep}{0pt}
     \setlength{\parsep}{0pt}
    \setlength{\topsep}{0pt}
    \setlength{\partopsep}{0pt}
    \setlength{\leftmargin}{2em}
    \setlength{\labelwidth}{1.5em}
    \setlength{\labelsep}{0.5em} } }

\newcommand{\squishend}{
  \end{list}  }

\newcommand{\betterparagraph}[1]{\noindent \textbf{#1. }}

\newcommand{\bench}{\textsc{XRBench}\xspace}

\newcommand{\benchscore}{\bench \textsc{score}\xspace}

\newcommand{\abbrvWorkload}{MTMM} 

\newcommand{\workload}{multi-task multi-model\xspace}

\usepackage[accepted]{mlsys2023}

\def\BibTeX{{\rm B\kern-.05em{\sc i\kern-.025em b}\kern-.08em
    T\kern-.1667em\lower.7ex\hbox{E}\kern-.125emX}}

\usepackage[firstpage]{draftwatermark}
\SetWatermarkText{
 \hspace*{4.5in}
 \raisebox{10in}{
  \includegraphics[height=0.9in]{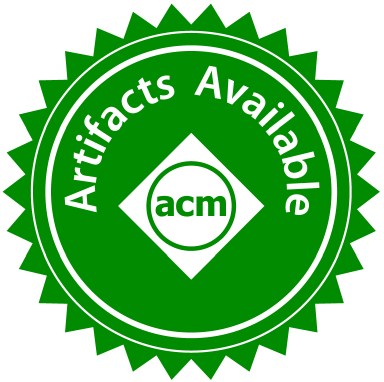}
  \includegraphics[height=0.9in]{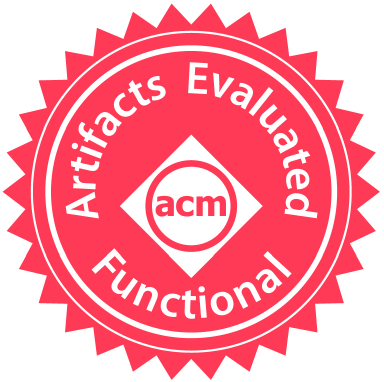}
 }
}
\SetWatermarkAngle{0}

\begin{document}

\twocolumn[

\mlsystitle{XRBench: An Extended Reality (XR) Machine Learning Benchmark Suite for the Metaverse}



\mlsyssetsymbol{equal}{*}

\begin{mlsysauthorlist}
\mlsysauthor{Hyoukjun Kwon}{uci,meta}
\mlsysauthor{Krishnakumar Nair}{meta}
\mlsysauthor{Jamin Seo}{gatech,equal}
\mlsysauthor{Jason Yik}{harvard,equal}\\
\mlsysauthor{Debabrata Mohapatra}{meta}
\mlsysauthor{Dongyuan Zhan}{meta}
\mlsysauthor{Jinook Song}{meta}
\mlsysauthor{Peter Capak}{meta}
\mlsysauthor{Peizhao Zhang}{meta}\\
\mlsysauthor{Peter Vajda}{meta}
\mlsysauthor{Colby Banbury}{harvard}
\mlsysauthor{Mark Mazumder}{harvard}
\mlsysauthor{Liangzhen Lai}{meta}
\mlsysauthor{Ashish Sirasao}{meta}\\
\mlsysauthor{Tushar Krishna}{gatech}
\mlsysauthor{Harshit Khaitan}{meta}
\mlsysauthor{Vikas Chandra}{meta}
\mlsysauthor{Vijay Janapa Reddi}{harvard}

\end{mlsysauthorlist}

\mlsysaffiliation{uci}{EECS, University of California, Irvine, Irvine, California, USA}
\mlsysaffiliation{meta}{Meta, Menlo Park, California, USA}
\mlsysaffiliation{gatech}{ECE, Georgia Institute of Technology, Atlanta, Georgia, USA}
\mlsysaffiliation{harvard}{SEAS, Harvard University, Cambridge, Massachusetts, USA}

\mlsyscorrespondingauthor{Hyoukjun Kwon}{hyoukjun.kwon@uci.edu}

\mlsyskeywords{XR, Benchmark, AR/VR, Metaverse}

\vskip 0.1in

\begin{abstract}

Real-time \workload (\abbrvWorkload{}) workloads, a new form of deep learning inference workloads, are emerging for applications areas like extended reality (XR) to support metaverse use cases. These workloads combine user interactivity with computationally complex machine learning (ML) activities. Compared to standard ML applications, these ML workloads present unique difficulties and constraints. Real-time \abbrvWorkload{} workloads impose heterogeneity and concurrency requirements on future ML systems and devices, necessitating the development of new capabilities. This paper begins with a discussion of the various characteristics of these real-time \abbrvWorkload{} ML workloads and presents an ontology for evaluating the performance of future ML hardware for XR systems. Next, we present \bench, a collection of \abbrvWorkload{} ML tasks, models, and usage scenarios that execute these models in three representative ways: cascaded, concurrent, and cascaded-concurrent for XR use cases. Finally, we emphasize the need for new metrics that capture the requirements properly. We hope that our work will stimulate research and lead to the development of a new generation of ML systems for XR use cases. \revision{XRBench is available as an open-source project: \url{https://github.com/XRBench}}
\vspace{-3mm}
\end{abstract}

]

\printAffiliationsAndNotice{\mlsysEqualContribution}


\thispagestyle{empty}
\pagestyle{empty}

\section{Introduction}
\label{sec:introduction}


Applications based on machine learning (ML) are becoming prevalent. The number of ML models that must be supported on the edge, mobile, and data centers is growing. The success of ML across tasks in vision and speech recognition is furthering the development of increasingly sophisticated use cases. For instance, the \textit{metaverse}~\cite{metaverse} combines multiple unit use cases (e.g., image classification and speech recognition) to create more sophisticated use cases such as real-time interactivity via virtual reality. 
Such sophisticated use cases demand more functionality, for which application engineers are increasingly relying on composability; rather than developing different large models for use cases, they are combining multiple smaller and specialized ML models to compose task functionality~\cite{barham2022pathways}. 



In this paper, we focus on this new class of ML workloads referred to as \workload (\abbrvWorkload{}) ML workloads, specifically in the context of extended reality (XR) for metaverse use cases. A real-time \abbrvWorkload{} application for extended reality is illustrated in \autoref{fig:Overview}. The figure depicts how several \abbrvWorkload{} models can be cascaded and operated concurrently, sometimes dynamically subject to certain conditions, to provide complex application-level functionality. The center section of the figure demonstrates that processing throughput requirements can vary depending on the usage scenario. The right side of the figure shows how there can be a variety of interleaved execution patterns for each of the concurrent jobs. \abbrvWorkload{} workloads exhibit model heterogeneity, expanded computation scheduling spaces~\cite{kwon2021heterogeneous}, and usage-dependent real-time constraints, which makes them challenging to support compared to today's  single-task single-model (STSM) workloads. 


\insertWideFigure{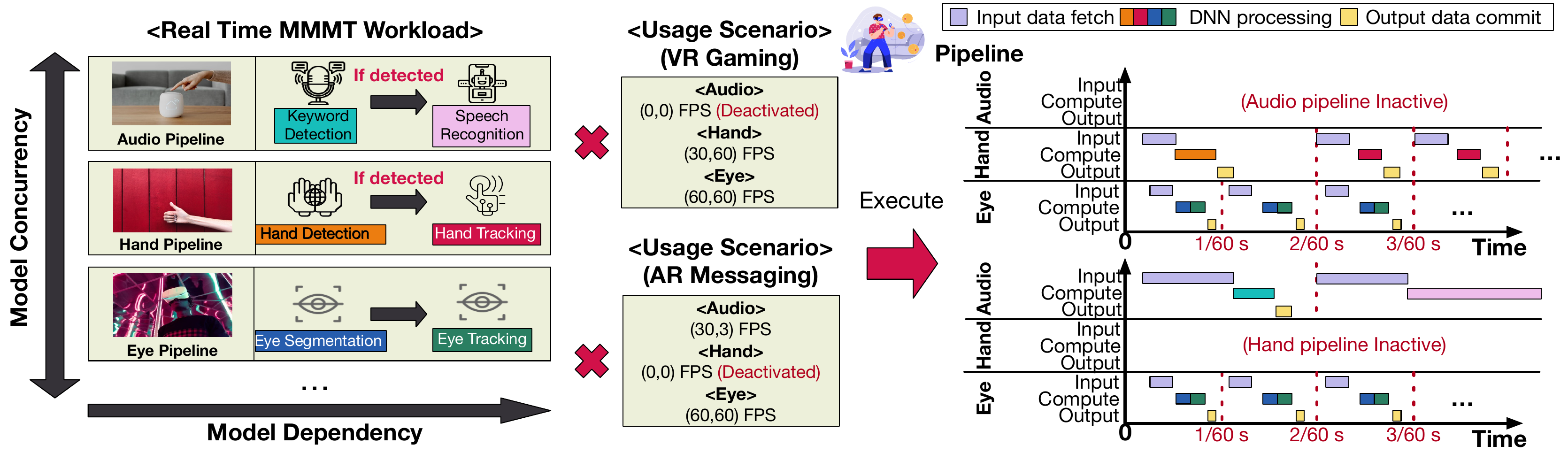}{An example real-time \workload (\abbrvWorkload{}) ML workload and an example execution timeline.
\vspace{-1.5mm}
}

We identify three key issues that arise with \abbrvWorkload{} workloads that present interesting system-level design challenges. The first is \emph{scenario-driven behavior}. All ML pipelines operate at a set frames per second (FPS) processing rate that is determined by a particular use case (e.g., virtual reality gaming, augmented reality social interaction, and outdoor activity recognition). A scenario may sometimes even demand zero FPS (i.e., deactivating a model) for models not required for the scenario. This fluctuating FPS is due to the context-based behavior that drives system resource utilization, which presents a hurdle when designing the underlying DNN accelerator---the heterogeneous workload makes it difficult to employ traditional DNN specialization.

Second, \abbrvWorkload{} workloads exhibit \emph{complex dependencies}. XR use cases display substantial {data dependency} (e.g., eye segmentation to tracking) and {control dependency} (e.g., hand detection to tracking) across models. These severe model-dependency limitations have ramifications for the underlying hardware and software scheduling space. 
In particular, the control flow dependencies make workload tasks dynamic, creating complexities for runtime scheduling.

Third, XR workloads have stringent user \emph{quality of experience (QoE) requirements.} A key distinguishing factor of \abbrvWorkload{} workloads from STSM ML workloads is the importance of understanding how to quantify the {aggregated QoE metric} across concurrent ML tasks at a system level. The resulting user QoE extends {beyond the computational performance (latency or throughput) of a single model}, which motivates the need for new metrics. Simple metrics like latency and/or FPS \textit{do not} capture the complex interactions of all these models across diverse scenarios. For example, the latency of each inference cannot be the absolute metric, since improving latency beyond the deadline set by the target processing rate may not improve the overall processing rate (e.g., the processing rate may be bound by the sensor input stream rather than inference time). Therefore, we need a new scoring metric that can capture the aggregate performance of the \abbrvWorkload{} workloads under different usage scenarios. The scoring metric must collectively consider all system aspects (model accuracy, achieved processing rate compared to the target processing rate, energy, etc.).

Collectively, not only do these three unique characteristics present system design challenges for XR, but they also make it challenging to benchmark and systematically characterize the performance of XR systems. Unfortunately, many of the characteristics and system-level concerns associated with \abbrvWorkload{} workloads are not fully understood. This is largely due to the lack of public knowledge regarding the realistic characteristics of \abbrvWorkload{} workloads derived from industry use cases. Consequently, the ML system design area for these workloads has yet to be explored. Furthermore, there is no benchmark suite of \abbrvWorkload{} workloads that reflects industrial use cases. Many industry and academic benchmark suites that exist today focus almost exclusively on STSM or \abbrvWorkload{} without cascaded models~\cite{reddi2020mlperf}. 


To address these deficiencies, we develop \bench, a real-time multi-model ML benchmark with new metrics tailored for real-time \abbrvWorkload{} workloads such as from the metaverse. 
\bench includes proxy workloads based on real-world industrial use cases taken from production scenarios. These proxy workloads encapsulate the end-to-end properties of \abbrvWorkload{} workloads at both the ML kernel and system levels, enabling the study of a vast design space. 

\bench includes scenario-based FPS requirements for ML use cases, which reflect the complex dependencies found in applications driving system-design research in a large organization invested in XR. It also presents representative QoE requirements for making system decisions. \bench consists of many usage scenarios of a metaverse end-user device that combines various unit ML models with different target processing rates to reflect the dynamicity and real-time features of \abbrvWorkload{} workloads. Furthermore, to enable comprehensive evaluations of ML systems using \bench, we also propose and evaluate new scoring metrics that encompass four distinct requirements for the QoE of real-time \abbrvWorkload{} applications: (1) the degree of deadline violations, (2) frame drop rate, (3) system energy consumption, and (4) model performance (e.g., accuracy).

\begin{table*}[t]
\centering
\setlength{\abovecaptionskip}{0pt}
\setlength{\belowcaptionskip}{0pt}
\scriptsize
\caption{\small \bench unit tasks and proxy unit models. 
Note that KD and SR are used for multiple task categories. Model performance requirements are 95\% of model performance (or, 105\% of error) reported in original papers, which opens the benchmark to various optimization techniques (e.g. mixed-precision), while ensuring reasonable prediction correctness. LT and GT refer to less than and greater than. For some models, we down-scale dataset resolution 
to adjust to the context of wearable/mobile devices, as we list in ~\autoref{sec:model_instances}.
}
\label{tab:unit_models}
\begin{tabular} {|@{~} l @{~}|l|l|l|l|l|}
\hline
\multicolumn{1}{|c|}{\textbf{Category}} &
\multicolumn{1}{c|}{\textbf{Task}} &
\multicolumn{1}{c|}{\textbf{Model}} &
\multicolumn{1}{c|}{\textbf{Dataset}} &
\multicolumn{1}{c|}{\textbf{Model Perf. Requirement}} 
\\
\hline

\hline
\multirow{5}{*}{\textbf{Interaction}} 
    & Hand Tracking (HT) & 
   Hand Shape/Pose~\cite{ge2019handshapepose} & 
   Stereo Hand Pose~\cite{stereohandpose}
   & AUC PCK, GT 0.948 \\
    \cline{2-5}
    & Eye Segmentation (ES) & RITNet~\cite{chaudhary2019ritnet} & OpenEDS 2019~\cite{garbin2019openeds} 
    & mIoU, GT 90.54  \\
    \cline{2-5}
    & Gaze Estimation (GE) & Eyecod~\cite{you2022eyecod} & OpenEDS 2020~\cite{palmero2021openeds2020} 
    & Angular Error, LT 3.39 \\
    \cline{2-5}
    & Keyword Detection (KD) & Key-Res-15~\cite{tang2018deep} & Google Speech Cmd~\cite{google_speech_commands} & Accuracy, GT 85.60 \\
    \cline{2-5}
    & Speech Recognition (SR) 
    & Emformer ~\cite{shi2021emformer} & LibriSpeech~\cite{panayotov2015librispeech}  & WER (others), LT 8.79 \\
    \cline{2-5}
\hline
\multirow{5}{1.8cm}{\textbf{Context Understanding}}
    & Semantic Segmentation (SS) & HRViT~\cite{gu2022multi} & Cityscape~\cite{cordts2016cityscapes} 
    & mIoU, GT 77.54 \\
    \cline{2-5}
    & Object Detection (OD) & D2Go~\cite{d2go} & COCO~\cite{lin2014microsoft} 
    & boxAP, GT 21.84 \\
    \cline{2-5}
    & Action Segmentation (AS) & TCN ~\cite{lea2017temporal} & GTEA~\cite{fathi2011learning} 
    & Accuracy, GT 60.8 \\
    \cline{2-5}
    & Keyword Detection (KD) & Key-Res-15~\cite{tang2018deep} & Google Speech Cmd~\cite{google_speech_commands} 
    & Accuracy, GT 85.60 \\
    \cline{2-5}
    & Speech Recognition (SR) 
    & Emformer ~\cite{shi2021emformer} & LibriSpeech~\cite{panayotov2015librispeech} 
    & WER (others), LT 8.79 \\
    \cline{2-5}    
\hline
\multirow{3}{*}{\textbf{World Locking}}
    & Depth Estimation (DE) & MiDaS~\cite{ranftl2020towards} & KITTI~\cite{geiger2012we}
    & $\delta>1.25$,LT 22.9 \\
    \cline{2-5}
    & Depth Refinement (DR) & Sparse-to-Dense~\cite{ma2018sparse} & KITTI~\cite{geiger2012we}
    & $\delta_1$, GT 85.5(100 samples) \\
    \cline{2-5}
    & Plane Detection (PD) & PlaneRCNN~\cite{liu2019planercnn} &
    KITTI~\cite{geiger2012we}
    & $AP^{0.6m}$, GT 0.37 \\
    \cline{2-5}
\hline
\end{tabular}
\vspace{-6mm}
\end{table*}

In summary, we make the following contributions:

\squishlist

\item We {provide a taxonomy of \abbrvWorkload{}-based workloads to articulate the unique features and challenges} of real-time workloads for metaverse use cases.

\item We present {\bench, an ML benchmark suite for real-time XR workloads}. We provide open-source reference implementations for each of the models to enable widespread adoption and usage. 

\item We {establish new scoring metrics for \bench that capture key requirements for real-time \abbrvWorkload{} applications} and conduct quantitative evaluations.

\item \revision{We make \bench available as an open-source project: \url{https://github.com/XRBench}}

\squishend


\section{\abbrvWorkload{} Workload Characteristics} 
\label{sec:\abbrvWorkload{}_characteristics}

\vspace{-1mm}


To assist XR systems research on real-time \abbrvWorkload{} workloads, we define a benchmark suite based on industrial metaverse \abbrvWorkload{} use cases. Before discussing the benchmark suite in~\autoref{sec:benchmark}, we first define the \abbrvWorkload{} classification and the characteristics of a realistic \abbrvWorkload{} workload, cascaded and concurrent \abbrvWorkload{}. 

\vspace{-2mm}

\subsection{Multi-model Machine Learning Workloads}
\label{sec:multi_model_workloads}

Unlike STSM workloads, \abbrvWorkload{} workloads include many models that lead to multiple model organization choices for constructing a workload instance. Based on the styles of those, we define three major classes:

\vspace{-2mm}

\squishlist
    {\item \textit{Cascaded \abbrvWorkload{} (cas-\abbrvWorkload{}):} Run multiple models back-to-back to enable one complex functionality (e.g., audio pipeline in~\autoref{fig:Overview})}.
    {\item \textit{Concurrent \abbrvWorkload{} (con-\abbrvWorkload{}):} Run multiple models independently at the same time to enable multiple unit functionalities (e.g., run Mask R-CNN~\cite{he2017mask} and PointNet~\cite{qi2018frustum} to perform 2D and 3D object detection during mapping and localization)}.
    {\item \textit{Cascaded and concurrent \abbrvWorkload{} (cascon-\abbrvWorkload{}):} Hybrid of cas- and con-\abbrvWorkload{}; connect multiple models back-to-back (cas-\abbrvWorkload{} style) to implement a complex ML pipeline and deploy multiple models (con-\abbrvWorkload{} style) for the other functionalities. (e.g., the VR gaming usage scenario in~\autoref{fig:Overview}).}
\squishend

\vspace{-2mm}

\textit{Static vs. Dynamic: }In addition to the model organization style, the model execution graph can be static or dynamic depending on the unit pipelines defined for a workload.
For example, as shown in~\autoref{fig:Overview}, hand tracking can be deactivated if the hand detection model detects no hand.

In recent applications that encompass extended reality, we can observe dynamic and real-time cascon-\abbrvWorkload{} style workloads~\cite{kwon2021heterogeneous}, which represent some of the most complicated ML inference workloads today.
Although such dynamic and real-time cascon-\abbrvWorkload{} style workloads are emerging, we lack a benchmark suite for dynamic cascon-\abbrvWorkload{} workloads. Consequently, there has been no deep understanding of the features and challenges from dynamic cascon-\abbrvWorkload{}, which we discuss next.
%

\vspace{-2mm}

\subsection{Dynamic Cascon-\abbrvWorkload{} Features and Challenges}
\label{sec:challenges}

Cascaded and concurrent \abbrvWorkload{} workloads are an emerging class of ML inference tasks. They have unique features and issues that do not exist in conventional ML workloads. We outline such aspects and analyze the issues of cascon-\abbrvWorkload{} workloads for metaverse (XR) applications.

\vspace{-3mm}

\subsubsection{Scenario-driven Workloads}
\label{subsec:usage_scenario}

\vspace{-1mm}

Metaverse workloads come from various different usage scenarios. A usage scenario refers to specific user experiences while utilizing a device or service. Gaming (e.g., VR gaming) and social (e.g., AR messaging) are example usage situations. Usage scenarios can be generated by combining several unit tasks, such as hand tracking or keyword detection. So, metaverse workloads must take the usage scenario into account to determine which unit tasks should be included, which is one of their distinctive elements compared to workloads in benchmarks such as MLPerf~\cite{reddi2020mlperf} and ILLIXR~\cite{huzaifaillixr}.

\vspace{-2mm}


\subsubsection{Real-Time Requirements}
\label{subsec:realtime}
\vspace{-1mm}

Many existing ML-based applications often employ a single model inference to input (e.g., image or text). In contrast, metaverse devices are frequently required to continually execute inferences of a set of models in order to provide continuous user experiences (e.g., a user plays a VR game for 1 hour). As inference runs contribute to user experiences, it is only reasonable that a strong quality of user-driven experience (QoE) is required. 
In the context of multi-model inference, QoE can be represented by
processing rate (i.e., inferences per second, such as FPS for models with frame-based inputs) or processing deadlines, hence introducing real-time processing requirements. Consequently, just as ML benchmarks must satisfy a certain level of accuracy for the quality of results~\cite{reddi2020mlperf}, XR benchmarks must also provide target processing rates.

\vspace{-2mm}

\subsubsection{Dynamic Cascading of Models}
\label{subsec:dynamic}
\vspace{-1mm}

Metaverse applications commonly utilize numerous models. For example, hand-based interaction capabilities can be enabled by cascaded hand detection and hand tracking models. Such models are often cascaded (i.e., run sequentially in a back-to-back manner), and such cascaded models are characterized as a pipeline of models (or an ML pipeline).

\autoref{fig:Overview} presents three ML pipeline examples. Such pipelines need to be transformed into data dependencies across models, which need to be considered while scheduling computations ~\cite{kwon2021heterogeneous}. \abbrvWorkload{} ML pipelines may deactivate one or more downstream models based on the upstream model's results. For instance, when no hand is detected, the hand tracking pipeline does not initiate the downstream hand tracking model. Such a dynamic aspect presents another problem for scheduling computation. In addition, it indicates that metaverse benchmarks must include different usage scenarios that reflect the dynamic nature of metaverse workloads.

\vspace{-2mm}
\subsubsection{Battery Life and Device Form Factor}
\label{subsec:baterylife}
\vspace{-1mm}

The wearable form factor of metaverse devices makes thermal tolerances and battery life first-order priorities for user experience. For example, if the heat dissipation is excessively high, it may lead to skin discomfort or burns. Long battery life is critical since wearable devices are intended to be used continuously throughout the day, but the form factor places further constraints on battery size, even compared to other edge devices. For example, a recent metaverse device~\cite{google_glasses_spec} has an 800~mAh battery, which is a fifth of the size of the battery in a modern mobile device (e.g., 4000~mAh in Samsung Galaxy S20~\cite{galaxy_s20_spec}). Energy consumption must be a primary optimization priority for metaverse end-user devices. All of the requirements (i.e., scenario-driven tasks, real-time requirements, dynamic cascading, battery life, and form factor) translate into energy constraints. So the benchmark needs to contain energy goals to ensure a device provides a good user experience.

\vspace{-3mm}

\section{\bench}
\label{sec:benchmark}

\vspace{-1mm}


Real-time cascon-\abbrvWorkload{} workloads for the metaverse are distinctive due to the discussed characteristics and obstacles. As a result, this domain necessitates a new method of defining benchmarks in comparison to traditional model-level benchmarks alone. In this section, we outline what we consider to be the most important characteristics of an \abbrvWorkload{} benchmark. Then, we describe \bench, the first benchmark of its kind for extended reality applications.

\vspace{-3mm}

\subsection{Benchmark Principles}

\vspace{-1mm}

To systematically guide the design of \abbrvWorkload{} benchmarks, we focus on the key requirements for such a benchmark:

\vspace{-2mm}

\squishlist
    {\item \textit{Usage Scenarios:} A set of real-world usage situations based on production use cases and a list of models to be run for each usage scenario must be defined.}
    {\item \textit{Model Dependency:} As certain ML models are cascaded, model dependencies across the task must be specified to study resource allocation and scheduling effects.}
    {\item \textit{Target Processing Rates:} Provide meaningful and applicable real-time requirements and processing rates for each model in each usage scenario to establish application behavior and system performance expectations.}
    {\item \textit{Variants of a Usage Scenarios:} To reflect the dynamic nature of model execution and enable apples-to-apples comparisons, the benchmark must give numerous scenarios with distinct active time windows for each model.}
\squishend

\vspace{-2mm}

Based on the requirements, we define \bench. We first discuss unit models and usage scenarios in \bench, then describe its evaluation infrastructure and scoring techniques. Later in \autoref{sec:eval}, we show why these principles are important by conducting architectural analysis using \bench.   

\vspace{-3mm}

\subsection{Unit-level ML Models}
\label{subsec:benchmark_models}

\vspace{-2mm}




Based on our experience in the metaverse (XR) domain, we define the first dynamic cascon-\abbrvWorkload{} benchmark that reflects metaverse use cases. 
There are three main task categories in \bench, listed in \autoref{tab:unit_models}: real-time user interaction, understanding user context, and world locking (AR object rendering on the scene). These categories are based on real-world industrial use cases for the metaverse. For each unit task, we choose a representative reference model from the public domain. 

\revision{When selecting models, we consider two aspects: (1) model performance (e.g., accuracy) and (2) efficiency (the number of FLOPs and parameters).}
Additionally, we list datasets and accuracy requirements for each unit task. More information for each unit task, including specific open-source model instances and dataset can be found in \autoref{sec:model_instances}. 

\begin{table*}[t]
\centering
\setlength{\abovecaptionskip}{0pt}
\setlength{\belowcaptionskip}{0pt}
\scriptsize
\caption{\small Target processing rates (FPS). \revision{Eye and speech pipelines have data (D) or control (C) dependencies.}}
\label{tab:usage_scenario_processing_rates}
\begin{tabular} {|@{~} l @{~}|l|c|c|c|c|c|c|c|c|c|c|c|}
\hline
\multirow{2}{*}{\textbf{Usage Scenario}} 
& \multirow{2}{*}{\textbf{HT}} 
& \multicolumn{2}{c|}{\textbf{Eye Pipeline}}
& \multicolumn{2}{c|}{\textbf{Speech Pipeline}}
& \multirow{2}{*}{\textbf{SS}}
& \multirow{2}{*}{\textbf{OS}}
& \multirow{2}{*}{\textbf{AS}}
& \multirow{2}{*}{\textbf{DE}}
& \multirow{2}{*}{\textbf{DR}}
&\multirow{2}{*}{\textbf{PD}}
& \multirow{2}{*}{
\textbf{Example Usage Scenario Description}
} \\
\cline{3-6}
&  
& \multicolumn{2}{c|}{\textbf{ES $\rightarrow$ GE (dep: D)}}
& \multicolumn{2}{c|}{\textbf{KD $\rightarrow$ SR (dep: C)}} 
& 
& 
& 
& 
& 
& 
& 
\\
\hline

\hline
Social Interaction A & 30 & 60 & 60&  &  &  &  &  &  &  30 &  & \multicolumn{1}{l|}{AR messaging with AR object rendering} \\
\hline
Social Interaction B & & 60 & 60 &   &  &  &  & 30 &  &  &  & \multicolumn{1}{l|}{In-person interaction with AR glasses}\\
\hline
Outdoor Activity A &  &  &  & 3 & 3 &  10 & 30 &  &  &  &  & \multicolumn{1}{l|}{Hiking with smart photo capture} \\
\hline
Outdoor Activity B &  &  &  & 3 & 3 & & 30 &  &  &  &  & \multicolumn{1}{l|}{Rest during hike} \\
\hline
AR Assistant  &  &  &  & 3 & 3 &  10 & 10 &  & 30 &   & 30 & \multicolumn{1}{l|}{Urban walk with informative AR objects} \\
\hline
AR Gaming  & 45 &  &  &  &  &  &  &  &   30 &  & 30 & \multicolumn{1}{l|}{Gaming with AR object} \\
\hline
VR Gaming  & 45 & 60 & 60 &  &  &  &  &  & &  & & \multicolumn{1}{l|}{Highly-interactive Immersive VR gaming}  \\
\hline
\end{tabular}
\vspace{-4mm}
\end{table*}






\vspace{-2.5mm}

\subsubsection{Interaction} 

\vspace{-1.5mm}

Real-time user interaction tasks enable users to control metaverse devices using various input methods, including hand movements, eye gaze, and voice inputs. Therefore, we include corresponding ML model pipelines: hand pipeline (end-to-end model performing hand detection and tracking), eye pipeline (ES and GE), and voice pipeline (KD and SR).

\vspace{-2.5mm}

\subsubsection{Context Understanding} 

\vspace{-1.5mm}

Context understanding tasks use multi- (e.g., VIO) or single-modal (e.g., audio) inputs to detect the context surrounding users so that a metaverse device can provide the appropriate user services. When a metaverse device detects that a user has entered a hiking trail, for example, it can provide the user with meteorological information. Context understanding models include scene understanding (SS, OD, and AS) and audio context understanding (KD and SR).

\vspace{-2mm}

\subsubsection{World Locking} 

\vspace{-2mm}

A metaverse device must comprehend distances to real-world surfaces and occlusions in order to depict an augmented reality (AR) object on the display. These tasks are handled by models in the world-locking category, which includes a depth estimate model, a depth refinement model, and a plane detection model. The depth model is used to calculate the correct size of augmented reality (AR) objects, while the plane detection model identifies real-world surfaces that can be used to depict metaverse objects.

\vspace{-3mm}

\subsection{Usage Scenarios and Target Processing Rates}
\label{subsec:usage_scenarios}

\vspace{-1mm}

The models in \autoref{tab:unit_models} are selectively active with varying target processing rates depending on usage scenarios, as explained in \autoref{subsec:usage_scenario}. For example, the user experience of an AR game based on intensive hand interaction requires high hand-tracking speeds. The speech pipeline may be completely stopped if the game does not use speech input. During outdoor activities like hiking, however, an AR-enabled metaverse device may not require hand-tracking functionality but must be prepared for user speech input.


To reflect the different usage scenarios and target processing rate characteristics, we chose five realistic metaverse scenarios: (1) social interaction (AR messaging with AR object rendering), (2) outdoor activity (smart photo capture during hiking), (3) AR assistant (AR information display based on user contexts), (4) AR gaming, and (5) VR gaming.

Even within the same usage scenario, active models can differ because of the dynamic nature of cascon-\abbrvWorkload{} workloads.
For example, in an outdoor activity (hiking) usage scenario, when a user takes a break and tries to utilize an AR device (e.g., navigation and photo capturing), the hand tracking model will be engaged, unlike the previous hiking scenario. Considering such variability within usage scenarios, we suggest two versions (A and B) of social interaction and outdoor activity scenarios. \autoref{tab:usage_scenario_processing_rates} describes the usage scenario variants. In addition, we specify a target processing rate for each model with three levels: High (60Hz or 45Hz), Medium (30Hz), and Low (10Hz). SR has a processing rates of 3Hz, which models the 320ms left context size utilized in its original paper~\cite{shi2021emformer}. We assign target processing speeds depending on the usage scenarios based on practical metaverse use cases. We suppose that a metaverse device identifies the active models and their processing rates (i.e., usage scenario) based on the specific application launched by a user.

\vspace{-3mm}

\subsection{Input Sources and Load Generation}
\label{subsec:input_sources}

\vspace{-1mm}

\insertWideFigure{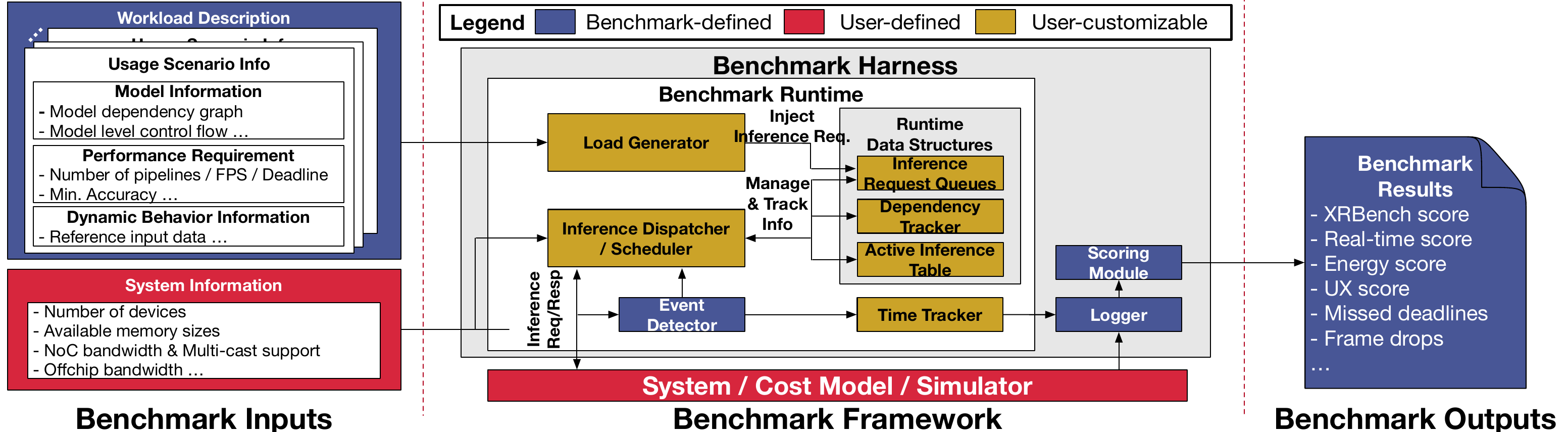}{An overview of the benchmark harness, \bench. 
\vspace{-2mm}
}

Metaverse devices utilize multiple sensors with varying modalities. To model the sensors, we use the settings listed in~\autoref{tab:input_sources} for the unit models  in~\autoref{tab:unit_models}. The camera is the input source of images used by computer vision models. The lidar sensor 
provides a sparse depth map to the depth refinement model 
using RGBd data. The microphone receives audio inputs for speech models (KD and SR).

\begin{table}[t]
\vspace{-2mm}
\scriptsize
\centering
\setlength{\abovecaptionskip}{0pt}
\caption{\small Three main input sources to a metaverse device. We align all the input streaming rates to be 60 FPS for multi-modal models (e.g., DR in~\autoref{tab:unit_models}). We also model jitters for each data frame.}
\label{tab:input_sources}
\begin{tabular} {|@{~} l @{~}|l|r|l|}
\hline
\multicolumn{1}{|c|}{\textbf{Input Source}} &
\multicolumn{1}{c|}{\textbf{Input Type}} &
\multicolumn{1}{c|}{\textbf{Streaming Rate}} &
\multicolumn{1}{c|}{\textbf{Jitter}} 
\\
\hline

\hline
Camera & Images & 60 FPS & $\pm$ 0.05 $ms$  \\
\hline
Lidar & Sparse Depth Points & 60 FPS & $\pm$ 0.05 $ms$ \\
\hline
Microphone & Audio & 3 FPS & $\pm$ 0.1 $ms$ \\

\hline
\end{tabular}
\vspace{-6mm}
\end{table}





The arrival time of the input data in an actual system can vary slightly from the projected time based on the streaming rate, depending on multiple circumstances (e.g., system bus congestion). In numerous research analyses, jitter is frequently disregarded. However, in genuine production usage circumstances, jitter might result in sporadic frame dropouts, which degrades QoE. To represent such aspects, we apply a jitter to each data frame, as shown in  \autoref{tab:input_sources}, and we alter the injection time of inference requests accordingly. 

\vspace{-3mm}

\subsection{Benchmark Harness}
\label{subsec:benchmark_design}

\vspace{-1mm}


A harness orchestrates the execution of the models, respecting the dependencies. We illustrate the structure of the benchmark harness in ~\autoref{fig:XRBenchToolFlow}. The harness takes workload and system information as input, and 
generates reports that contain not only the scores (overall score and its break-downs; to be discussed in~\autoref{sec:metric}) but also detailed performance statistics such as the amount of delay over deadline, frame drop, execution timeline, and so on. We include this detailed information in the reports to help users use \bench to guide their system designs.

The harness consists of a runtime, logger, and scoring module. The runtime contains a load generator which intermittently generates jittered inference requests. The inference dispatcher/scheduler is the core component of the runtime, which (1) selects the next inference requests to be dispatched when a hardware entity (e.g., accelerator) becomes available, (2) tracks the model and frame dependencies, and (3) dispatches inferences to the machine learning system to be evaluated (which may be a real physical system, analytical cost model, or simulator). The runtime components include an event detector, score tracker, and various data structures (request queue, active inference table, dependency table, etc.) that assist the dispatcher and scheduler.

\bench requires users to finish a certain number of runs, which equals to the target processing rate within a set duration (default: one second), to ensure the real-time requirement is satisfied. \bench provides a simple latency-greedy (for cost model or simulator-based runs) or a round-robin style scheduler (for real systems) for models within each usage scenario. Users can replace the scheduler and other components highlighted in yellow boxes in~\autoref{fig:XRBenchToolFlow} to model their runtime or system software's behavior. Much like in a traditional ML system or ML benchmark~\cite{reddi2020mlperf}, optimizing the software stack is crucial to the hardware's success, and \bench encourages such optimizations.




\vspace{-3mm}

\subsection{Deep Dive Example}
\label{subsec:benchmark_example}

\vspace{-1.5mm}

To clarify the roles of each piece in \bench, \autoref{fig:ExampleTimeline} provides an example execution timeline for the ``Social Interaction A'' usage scenario in~\autoref{tab:usage_scenario_processing_rates}. The execution graph on the left shows the active models, their processing rates, and the model dependencies.
The right side corresponds to a sample execution timeline for the scenario. The top-most section represents the input streaming from relevant input sources listed in~\autoref{tab:input_sources}. Each input source can have different initial delays and jitter. 

A compute engine (such as an accelerator) can only execute model inferences if it has access to the input data. In this example, we model a simple scheduler assuming that inferences can only begin if the input data is ready. Consequently, Eye Segmentation (ES) and Gaze Estimation (GE) for frame 0 begin once the input data retrieval for image frame 0 concludes. Additionally, GE runs after ES to satisfy their dependency. The multi-modal Depth Refinement (DR) model executes after image and depth point inputs are received. As DR's target processing rate is 30 FPS while depth point input  streams at 60 FPS, only every other input is used. As Hand Tracking (HT) also operates at 30 FPS, it skips every other image frame. 
The DR output is used for display-targeted AR object rendering. Therefore, DR results must be delivered by a certain time, which is a 30 FPS deadline (e.g., 2/60 s for frame 0) in this example.

\insertWideFigure{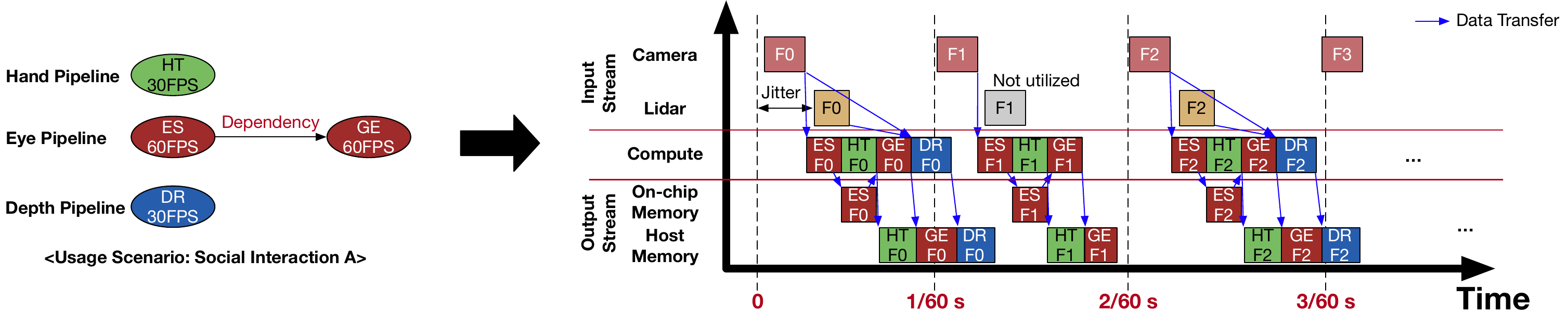}{An example execution timeline based on the Social Interaction A usage scenario in~\autoref{tab:usage_scenario_processing_rates}, \revision{FN refers to the frame N.}
\vspace{0mm}
}

In the execution timeline, the usage scenario is effectively supported and desired processing rates are attained.
However, ET and HT results are delivered past their desired deadlines ($1/60 \times frame$ and $1/30 \times frame$ seconds, respectively). This suggests that HT and ET latency must be reduced further in this example. Not always should latency itself be used as an optimization target, though, as 
latency reduction beyond deadlines may not improve the user experience. Even zero-latency inferences cannot increase the effective processing rate of a task beyond the input data streaming rates because future data cannot be processed without them. This raises the question: How should we quantify the performance of a real-time \abbrvWorkload{} system by taking into account the actual quality of the results for users? In the next subsection, we discuss new metrics that encompass the requirements and characteristics of XR tasks.

\vspace{-3mm}

\subsection{Scoring Metric\revision{s}}
\label{sec:metric}

\vspace{-1.5mm}

In~\autoref{subsec:benchmark_example}, we showed that evaluating a system for real-time \abbrvWorkload{} workloads is not trivial using the example in~\autoref{fig:ExampleTimeline}. \revision{For example}, lower inference latency does not always improve user experiences if the processing rate of each task is bound by the input data streaming rate. We need to capture such aspects when we evaluate a system for real-time XR workloads. Therefore, we define a new scoring metric \revision{, \benchscore,} considering all the aspects we discussed and propose it to be used as the overall performance metric in \bench. 

Based on the unique features of XR workload, we list the following score requirements for the benchmark: 

\vspace{-3mm}

\squishlist
{\item \textit{[Real-time]} The score should include a penalty if the latency exceeds the usage scenario's required performance constraints (i.e., missed deadlines).}
{\item \textit{[Low-energy]} The score should prioritize low-power designs as metaverse devices are energy-constrained.} 
{\item \textit{[Model quality]} The score should capture the  
 output quality delivered to a user from running all the models.}
{\item \textit{[QoE requirement]} The score should include a penalty if the FPS drops below the target FPS to maintain QoE.}
\squishend

\vspace{-2mm}


\insertFigure{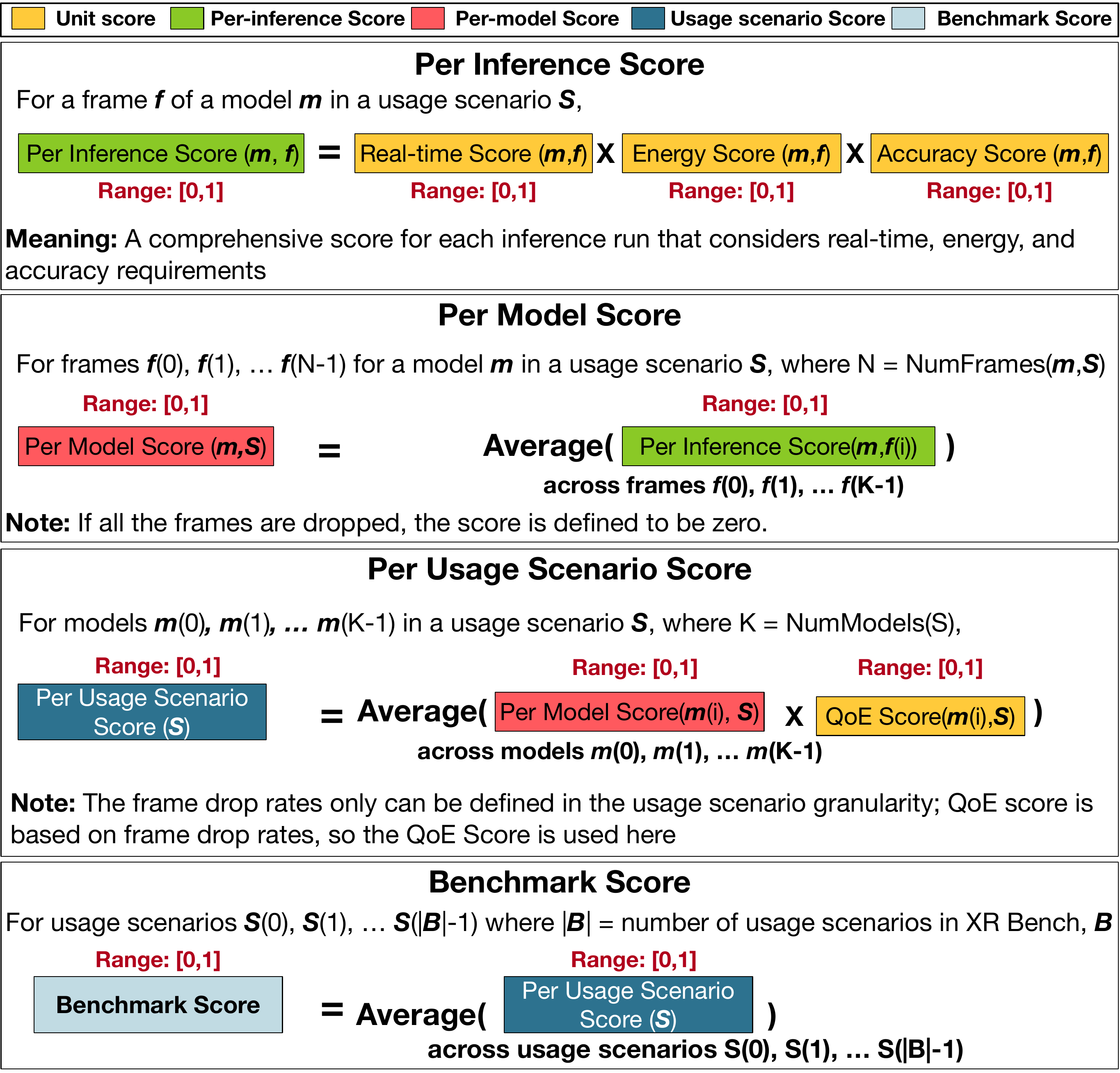}{A high-level overview of how we define benchmark scores at inference run, model, and usage scenario granularity using unit scores (real-time, energy, accuracy, and QoE scores).
\vspace{-0mm}
}

\begin{table}[t]
\vspace{-4mm}
\centering
\setlength{\abovecaptionskip}{0pt}
\setlength{\belowcaptionskip}{0pt}
\scriptsize
\caption{\small Symbols used in the formulation. (Only listing those not defined in~\autoref{box:base_defs} and ~\autoref{box:score_metrics}. 
\vspace{1mm}
}
\label{tab:symbol_table}
\begin{tabular} {|@{~} l @{~}|l|}
\hline
\multicolumn{1}{|c|}{\textbf{Symbol}} &
\multicolumn{1}{c|}{\textbf{Definition}}
\\
\hline
$M_{ID}$
& Model descriptor (model name)
\\
\hline
$inSrc_{ID}$
& Input source descriptor (e.g., sensor)
\\
\hline
$DS_{ID}$
& Dataset descriptor
\\
\hline
$QM_{ID}$
& Model quality metric descriptor (e.g., accuracy)
\\
\hline
$QM_{targ}$
& Target value of a model quality metric
\\ 
\hline
$QM_{Type}$
& The type of QM. Either Higher- or lower-is-better (HiB/LiB) 
\\
\hline
$Jt$
& Max absolute jitter in ms ($Jt \geq 0$)
\\ 
\hline
$Dep_{\mu}$
& A set of models on which model $\mu$ depend
\\
\hline
$L_{init}$
& Initialization latency (ms) of an input stream
\\ 
\hline
$L_{Inf}$
& Latency for an inference run
\\ 
\hline
$T_{start}(h,\mu)$
& Start time of an inference of $\mu$ on hardware $h$
\\ 
\hline
$T_{end}(h,\mu)$
& Completion time of an inference of $\mu$ on hardware $h$
\\ 
\hline
$HiB / LiB$
& Higher-is-better and lower-is-better metrics
\\ 
\hline
\end{tabular}
\vspace{-6mm}
\end{table}

We define four unit scores: real-time (RT), energy, accuracy, and QoE scores. Each score is constrained to be in the [0, 1] range for easy breakdown analysis. We multiply unit scores to reflect all of their aspects while keeping the results in the [0,1] range. We utilize averages to summarize scores for multiple inference runs on different frames for a model, multiple models within a usage scenario, etc. 
\revision{We focus on high-level intuitions with detailed formal definitions presented in ~\autoref{tab:symbol_table}, ~\autoref{box:base_defs}, and ~\autoref{box:score_metrics}.}


\begin{BOX}[ht]
\centering

\fbox{
 \addtolength{\linewidth}{-2\fboxsep}%
 \addtolength{\linewidth}{-2\fboxrule}%
\begin{minipage}{\linewidth}
\insertFormula{System/Benchmark Parameters}
{-1.6}
{
M_{ID}, inSrc_{ID}, DS_{ID}, QM_{ID} &\in str \\
FPS_{sensor}, FPS_{model}, InFrame_{ID} &\in \mathbb{N} \\ 
L_{init}, L_{inf}, Jt, QM_{targ}, T_{req} ,\epsilon &\in \mathbb{R} \\
QM_{Type} &= HiB ~|~ LiB
}
\insertFormula{Input Data Stream ($St_{input}$)}
{-1.6}
{
St_{input} = \{\sigma ~|~ \sigma = (inSrc_{ID}, FPS_{sensor}, L_{init}, Jt)\}
}
\insertFormula{Model Quality Goal ($Q$)}
{-1.6}
{
Q = (QM_{ID}, QM_{Targ}, QM_{Type})
}
\insertFormula{Unit Models ($M$)}
{-1.6}
{
M = \{\mu ~|~ \mu \in (M_{ID}, DS_{ID}, \sigma, Q) \wedge \sigma \in St_{input} \}
}
\insertFormula{Usage Scenario ($\theta$)}
{-1.6}
{
\theta = \{(\mu, Dep_{\mu}, FPS_{model})~|~\mu \in M \wedge Dep_{\mu} \subset M  \} 
}
\insertFormula{Benchmark Suite ($\Omega$)}
{-1.6}
{
\Omega = \{\theta_{1}, \theta_{2}, ... \theta_{NumScn}\} 
}
\insertFormula{Inference Request ($IR$)}
{-1.6}
{
IR = (\mu, InFrame_{ID})
}
\insertFormula{Inference Request Time($T_{req}(IR)$)}
{-1.6}
{
T_{req}(IR) = L_{init}(inSrc_{ID}) + \frac{InFrame_{ID}}{FPS_{Sensor}(inSrc_{ID})}  \\ 
+ 2 Jt ~(Dist(rand(inSrc_{ID}\times InFrame_{ID})) - 0.5) \\
where ~Dist(x) \in [0,1]~ \wedge ~x \in \mathbb{R}
}
\insertFormula{Inference Deadline($T_{dl}(IR)$)}
{-1.6}
{
T_{dl}(IR)= L_{init}(inSrc_{ID}) + \frac{InFrame_{ID} + 1}{SR(inSrc_{ID})}
}
\insertFormula{Inference Slack($T_{sl}(IR)$)}
{-2.5}
{
T_{sl}(IR)= T_{dl}(IR) - T_{req}(IR)
}
\end{minipage}
}
\vspace{-3mm}
\caption{Base Definitions}
\label{box:base_defs}
\vspace{-8mm}
\end{BOX}

\begin{BOX}[ht]
\centering

\fbox{
 \addtolength{\linewidth}{-2\fboxsep}%
 \addtolength{\linewidth}{-2\fboxrule}%
\begin{minipage}{\linewidth}
\insertFormula{Unit Score: Realtime Score ($RtScore(IR)$)}
{-1}
{
RtScore(IR) = \frac{1}{1+e^{k(L_{Inf}(IR)-T_{sl}(IR))}}
}
\insertFormula{Unit Score: Energy Score ($EnScore(IR)$)}
{-1}
{
EnScore(IR) = \frac{En_{max}-En(IR)}{En_{max}}
}
\insertFormula{Unit Score: Accuracy Score ($AccScore(IR)$)}
{-1.5}
{
AccScore(IR) =& max(1, rawAccScore(IR)) \\ 
rawAccScore(IR) =& 
    \begin{cases}
      \frac{QM_{measured}}{QM_{targ}},&\text{if}\ QM_{Type} = HiB\\
      \frac{QM_{targ}}{QM_{measured}+\epsilon},& \text{otherwise}
    \end{cases}\\
    & \text{where } \epsilon > 0 \wedge \epsilon \ll 1 \wedge \epsilon \in \mathbb{R}
}
\insertFormula{Unit Score: QoE Score ($QoEScore(\mu)$)}
{-1}
{
QoEScore(\mu) = \frac{NumFrm_{exec}(\mu)}{NumFrm(\mu)}
}
\insertFormula{Aggregated Score: Inference-wise Score ($Score_{inf}(IR)$)}
{-1.5}
{
Score_{inf}(IR) = &RtScore(IR) \times EnScore(IR) \\  
&\times AccScore(IR)
}
\insertFormula{Aggregated Score: Usage Scenario Score ($Score_{scn}(\theta)$)}
{-1}
{
Score_{scn}(\theta) =
\sum_{j=1}^{NumFrm(\mu)}{} 
\frac{Score_{inf}(IR)\times QoEScore(\mu)}{NumFrm(\mu) \times |\theta|}
}
\insertFormula{Aggregated Score: XRBench Score ($Score_{bench}$)}
{-2}
{
Score_{bench} = \frac{\sum_{\theta \in \Omega}{Score_{scn}(\theta)}}{|\Omega|}
}
\end{minipage}
}
\vspace{-4mm}
\caption{Score metrics}
\label{box:score_metrics}
\vspace{-5mm}
\end{BOX}

To model real-time requirements, we consider the following observations: (1) too much optimization on inference latency beyond the deadline does not lead to higher processing rates.
(2) reduced latency can still be helpful for scheduling other models. (3) violated deadlines gradually disrupt the user experience (e.g., Achieving 59 FPS for an eye-tracking model targeting 60 FPS won't significantly affect the user experience). Based on these observations, we search for a function that (1) gradually rewards/penalizes for reduced/increased latency near a deadline and (2) outputs 0 and 1 if the latency is well beyond (e.g., 0.5ms for a deadline of 10ms) or within the deadline , respectively. We find such a function by modifying the sigmoid function, which is widely used in ML models.

For energy, a lower-is-better metric, a naive way to compute energy score is computing the inverse of the energy consumption (example unit: 1/mJ). However, the range of the naive metric is unbounded, which makes it hard for component-wise analysis when it is combined with other scores bound in [0,1] ranges. Therefore, to bound the energy score within the [0,1] range as well, we utilize a large energy limit $E_{max}$ to define the top-end of the score. 

For accuracy score, we quantify how much the output correctness differs from the desired level using model-specific performance metrics (e.g., accuracy for classification, mIoU for segmentation, PCK AUC for hand tracking, etc.). Although there are many different metrics other than accuracy, we use the term, accuracy score, for simplicity.

Finally, we construct the \revision{\benchscore} in a hierarchical manner. \autoref{fig:ScoresOverview} illustrates how we combine scores along stages (unit, per-inference, per-model, per-usage scenario) to finally generate the overall \revision{\benchscore}.
We first compute the per-inference score by multiplying real-time, energy, and accuracy scores. The QoE score is not applied here as the frame drop rate only can be defined at the usage scenario level since the FPS requirements change depending on the usage scenario. Using the per-inference score, we construct the per-model score by computing the average across all processed frames. We do not include dropped frames since they will be considered in the QoE score. To compute the per-usage scenario score, we compute the average of the product of per-model score and QoE score across all the models within a usage scenario. 

\revision{Based on our approach discussed in this section, we formalize our score metrics in 
~\autoref{tab:symbol_table}, ~\autoref{box:base_defs}, ~\autoref{box:score_metrics}. We also provide more details in~\autoref{sec:appendix_formulation}.}



\revision{\bench reveals all individual scores to users to facilitate Pareto frontier analysis, in addition to \benchscore. In some cases, the industry may not wish to share the detailed performance breakdown of their system. Therefore, reporting breakdown scores is optional for \bench, while the overall \benchscore is mandatory. The released benchmark harness contains implementations of all scoring metrics.}



\vspace{-2mm}

\subsection{\revision{Limitations}}

\vspace{-1mm}


\revision{\bench focuses on ML components of XR workloads and does not model pre- and post-processing of inputs and outputs of ML pipelines. Such an approach is motivated by the significance of the ML processing time in XR systems.
}

\vspace{-3mm}
\section{Evaluation}
\label{sec:eval}

\vspace{-1mm}


In this section, we focus on three key questions to ascertain the value of \bench: (1) why the comprehensive overall score is necessary for the proper evaluation of XR tasks,
(2) why it is important to study the different usage scenarios that are included in \bench, and (3) what are the hardware implications of the \abbrvWorkload{} characteristics found in XR.

\vspace{-2mm}

\subsection{Methodology}
\label{subsec:eval_methodology}

\vspace{-1mm}

Metaverse applications run on wearable devices and the compute requirement for the workloads is heavy (tens for FPS requirements for multiple models). Therefore, considering the  capabilities of state-of-the-art mobile SoCs (e.g., 26 TOPS on Qualcomm Snap Dragon 888~\cite{qualcomm_sd888}), we model wearable devices with DNN inference accelerators that employ 4K and 8K PEs with 256 GB/s on-chip bandwidth and 8MiB of on-chip shared memory running at 1 GHz clock, similar to Herald~\cite{kwon2021heterogeneous}. 

\revision{\textbf{Simulated HW Accelerators.}}
\autoref{tab:eval_accelerators} shows \revision{various accelerator instances we evalute} in three accelerator styles: \textbf{FDA} (fixed-dataflow accelerator), \textbf{S}caled-out multi-\textbf{FDA} (two accelerator instances with the same dataflow style motivated by ~\cite{baek2020multi}), and \textbf{HDA} (heterogeneous dataflow accelerator)~\cite{kwon2021heterogeneous}. Depending on the style, we partition the 4K and 8K PEs into 2 or 4 accelerator instances. The WS (weight stationary) dataflow is inspired by NVDLA~\cite{nvdla} that parallelizes the output and input channels with input columns. OS (output-stationary) is a hand-optimized dataflow that parallelizes output rows and columns with a 16-way adder tree reducing input channel-wise partial sums. The RS (row stationary) dataflow is inspired by Eyeriss~\cite{chen2016eyeriss_isca} that parallelizes output channels, output rows, and kernel rows. \revision{Note that each accelerator in \autoref{tab:eval_accelerators} refers to an instance of hardware accelerator that can run \bench.}

\revision{\textbf{Simulation Methodology.} }We implement the framework illustrated in~\autoref{fig:XRBenchToolFlow} and plug in  MAESTRO~\cite{kwon2019understanding} as the analytical cost model to perform the different case studies. All the models are the same across the hardware platforms (8bit-quantized without other optimizations) and satisfy the accuracy goals (i.e., accuracy score = 1). 

\revision{\textbf{Modeling Dynamic Cascading.} }
To model the dynamic cascading between keyword detection and speech recognition, we apply pre-defined probabilities of user keyword utterances to corresponding usage scenarios (Outdoor A, Outdoor B, and AR Assistant). For outdoor activity scenarios, we apply 0.2 as the interaction is expected to be in a low frequency for the scenarios. For AR assistant, we apply 0.5 as the speech is the standard interaction method for the use case. For eye segmentation and gaze estimation pipeline, we first apply the probability of 1.0 to model pure data dependency and sweep the probability for a separate deep dive (\autoref{fig:DynamicCascading}).

\begin{table}[t!]
\scriptsize
\centering
\caption{Accelerator styles. Partitioning indicate the PEs to be deployed for each accelerator instance for SFDA and HDA.}
\label{tab:eval_accelerators}
\scriptsize
\begin{tabular}{|c|c|c|}
\hline
\textbf{Acc. ID}
& \textbf{Acc. Style}
& \textbf{Dataflow} \\
\hline
A
& \multirow{3}{*}{FDA}
& WS \\
  \cline{1-1} 
  \cline{3-3} 
  B & & OS \\
  \cline{1-1} 
  \cline{3-3}
  C & & RS \\
\hline
D &
\multirow{6}{*}{SFDA\footnote{~\cite{baek2020multi}}}
  & WS + WS (1:1 partitioning) \\
  \cline{1-1} 
  \cline{3-3} 
  E & & OS + OS (1:1 partitioning) \\
  \cline{1-1} 
  \cline{3-3} 
  F & & RS + RS (1:1 partitioning) \\
  \cline{1-1} 
  \cline{3-3}   
  G & & WS + WS + WS + WS (1:1:1:1 partitioning) \\
  \cline{1-1} 
  \cline{3-3} 
  H & & OS + OS + OS + OS (1:1:1:1 partitioning) \\
  \cline{1-1} 
  \cline{3-3} 
  I & & RS + RS + RS + RS (1:1:1:1 partitioning) \\  
\hline
%
%
J &
\multirow{4}{*}{HDA}
  & WS + OS (1:1 partitioning) \\
  \cline{1-1} 
  \cline{3-3} 
  K & & WS + OS (3:1 partitioning) \\
  \cline{1-1} 
  \cline{3-3} 
  L & & WS + OS (1:3 partitioning) \\
  \cline{1-1} 
  \cline{3-3}
  M & &  WS + OS + WS + OS (1:1:1:1 partitioning) \\
  \cline{1-1} 
  \cline{3-3} 
\hline
\end{tabular}
\vspace{-8mm}
\end{table}

\insertWideFigure{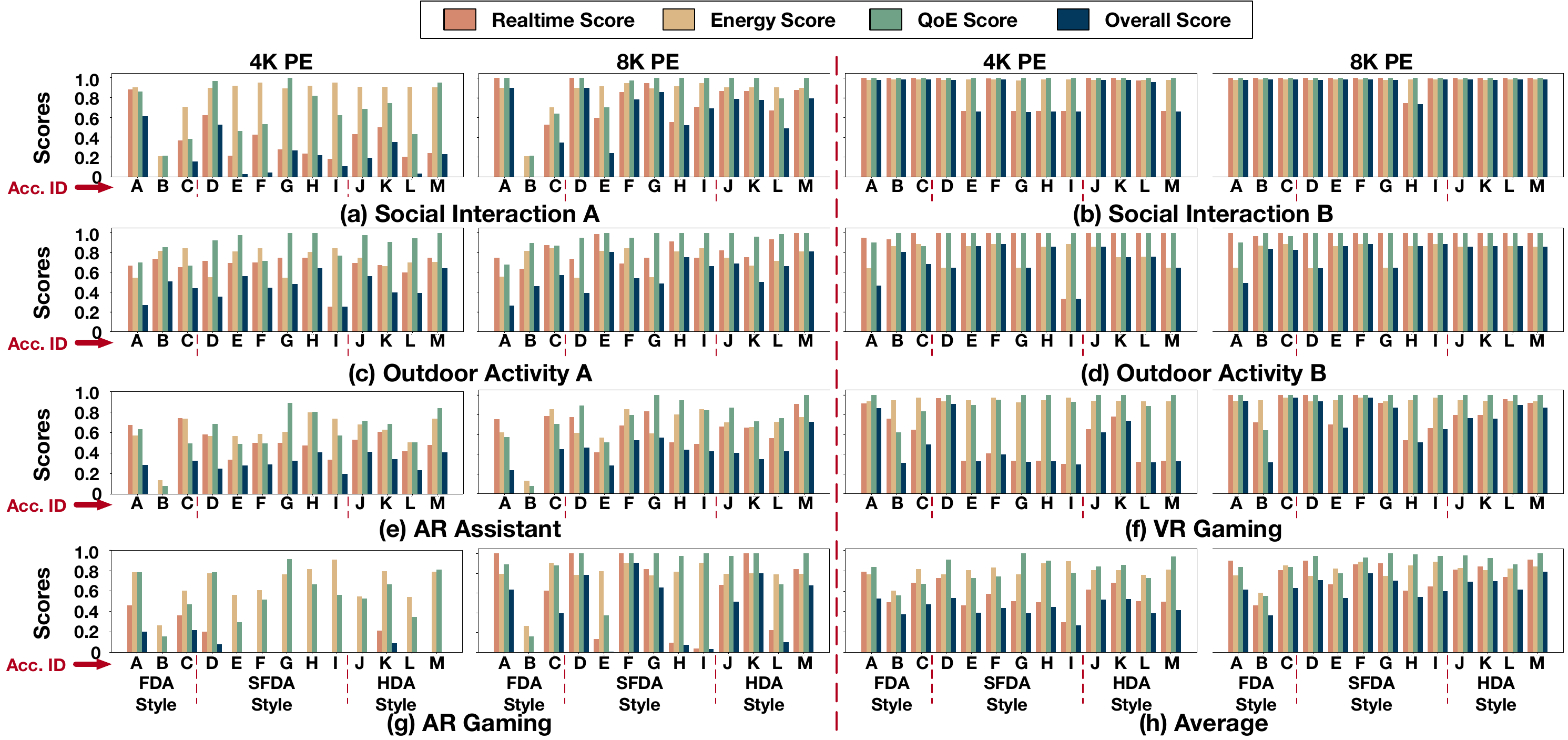}{The scores computed for each style of an accelerator system with 4K and 8K PEs. (a-g) the score break-downs for each usage scenario. (h) the average across scenarios. \revision{Overall score refers to \benchscore.}}

\insertFigure{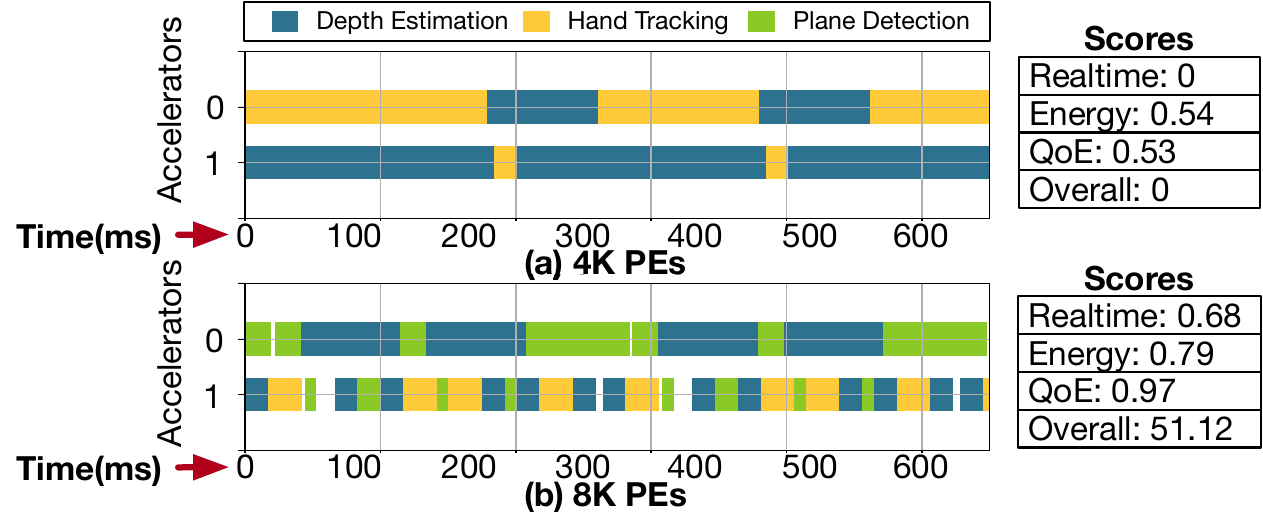}{Execution timeline of AR gaming scenario on 4k and 8k PE versions of WS and OS HDA accelerator (accelerator J).}



\subsection{Why the \revision{\benchscore} is a Necessary Metric}
\label{subsec:eval_score}


The intent of this section is to show that the overall scoring metric we present (Section~\ref{sec:metric}) is necessary for systematically evaluating XR systems. We present our evaluation results in~\autoref{fig:EvalResultsScore4K8K}, which shows score break-downs for each accelerator style running each usage scenario. 

\subsubsection{Overall Score Enables Comprehensive Analysis}

The real-time score quantifies the degree of deadline violation.
Higher-is-better for the real-time score; however, a high real-time score itself does not guarantee ideal system performance. For example, accelerator A with 8K PEs running the Outdoor Activity B (\autoref{fig:EvalResultsScore4K8K}, (d)) has a real-time score of 1.0, which indicates that most of the deadlines are met within a small margin. However, accelerator A misses 10.0\% of the frames (not shown) and has high energy consumption, 34.1\% greater than the most energy-efficient design (accelerator C).
Our scoring metric incorporates all aspects, including QoE score for frame drops and energy score for energy consumption, and it reports an overall score of 0.49, which is 42.9\% less than the best accelerator (I). 

As another example, for the AR Gaming scenario (\autoref{fig:EvalResultsScore4K8K}, (g)) on a 4K accelerator system, accelerator G achieved the greatest QoE score of 0.91 and a strong energy score of 0.76. However, its real-time score is zero due to heavily missed deadlines. That is, while the frame rate is overall close to the target as captured in the QoE score), a user will experience heavy output lag, which degrades the real-time experience. The real-time score captured this and led the overall score for this accelerator to be zero.


\subsubsection{Hardware Utilization is the Wrong Metric}


Hardware utilization is often used as a key metric for accelerator workloads since it can be directly translated to accelerator performance by multiplying utilization by the peak performance of the accelerator. However, we do not consider hardware utilization to be the right metric for real-time \abbrvWorkload{} applications, and as such we do not include it in the overall scoring metric (Section~\ref{sec:metric}).

Utilization does not consider frame drops or periodic workload injection. For example, \autoref{fig:timeline_ar_gaming_dual_WS_OS} shows the execution timelines for the 4K and 8K PE versions of accelerator J.
The 8K PE timeline (\autoref{fig:timeline_ar_gaming_dual_WS_OS}, (b)) has more gaps than the 4K PE timeline, which means the overall accelerator utilization is lower than that of the 4K PE accelerator (\autoref{fig:timeline_ar_gaming_dual_WS_OS}, (a)), making it seem as though the 4K PE accelerator is a better choice. However, the 4K-PE accelerator drops 47.1\% of the frames and completely fails to run the PD model, whereas the 8K-PE accelerator drops only 2.3\% of frames. 

Unlike the utilization alone, our score metrics properly capture the real-time requirement and QoE aspects. The frame drop rates of the 4K and 8K PE accelerators are captured in the QoE scores of 0.53 and 0.97, respectively. In addition, the large amount of deadline violations for the PD model in the 4K PE accelerator results in a real-time score of 0. Combining those unit scores into the overall score (\benchscore), we observe the scores of 0 and 0.51, which provides a better intuition to the comprehensive performance of an XR system considering all the considerations, including real-time requirements and QoE.



\insertFigure{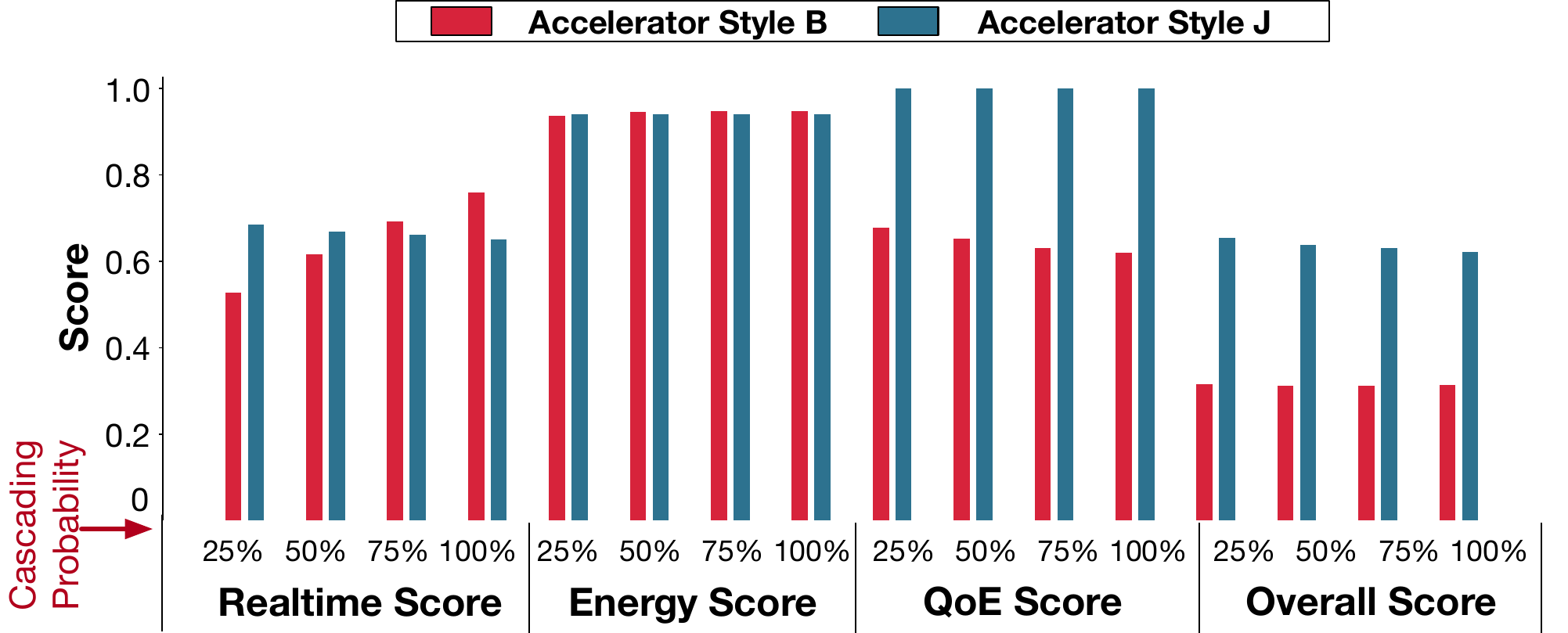}{Evaluated scores on accelerators B and J with 4K PEs running VR gaming scenario. We vary the probability of triggering GE after ES, modeling the dynamic cascading.}


\begin{table*}[t]
  \begin{threeparttable}[b]
\centering
\setlength{\abovecaptionskip}{0pt}
\setlength{\belowcaptionskip}{0pt}
\scriptsize
\caption{\small List of existing benchmarks related to ML and XR workloads, with comparison of workload characteristics and score metrics. $\triangle$ means the property is partially supported by the benchmark.}

\label{tab:related_works}\
\begin{tabular} {|c c|c|c|c|c|l|c|c|c|c|}
\hline 
\multicolumn{2}{|c|}{\multirow{4}{*}{\textbf{Benchmark}} }
    & \multicolumn{5}{c|}{\textbf{Workload Characterisitcs}}
    & \multicolumn{4}{c|}{\textbf{Score Metrics}} \\
    \cline{3-11}
& 
    & \multicolumn{1}{p{1.2cm}|}{\centering\textbf{Cascon-MMMT}} 
    & \multicolumn{1}{p{1.1cm}|}{{\multirow{2}{1.1cm}{\centering\textbf{Dynamic Workload}}}} 
    & \multicolumn{1}{p{1.1cm}|}{{\multirow{2}{1.1cm}{\centering\textbf{Real-time Scenarios}}}} 
    & \multicolumn{1}{p{0.7cm}|}{{\multirow{2}{0.7cm}{\centering\textbf{Focus on ML}}}}
    & \multicolumn{1}{p{1.1cm}|}{{\multirow{2}{1.1cm}{\centering\textbf{Device Scope}}}} 
    & \multicolumn{1}{c|}{\multirow{2}{*}{\centering\textbf{Latency}} }
    & \multicolumn{1}{c|}{\multirow{2}{*}{\centering\textbf{Energy}}} 
    & \multicolumn{1}{p{1.5cm}|}{\multirow{2}{1.5cm}{\centering\textbf{Accuracy
    }} }
    & \multicolumn{1}{c|}{\multirow{2}{*}{\centering\textbf{QoE}} } \\
\hline
\multicolumn{1}{|c}{\multirow{8}{*}{\textbf{ML}}}
    & \multicolumn{1}{|l|}{\shortstack[l]{MLPerf Inference\tnote{a}}} 
    &  
    &  
    & \checkmark 
    & \checkmark 
    & server 
    & \checkmark 
    & \checkmark 
    & \checkmark  
    & \\
    \cline{2-11}
    &  \multicolumn{1}{|l|}{\shortstack[l]{MLPerf Tiny\tnote{b} }} 
    &
    &
    & \checkmark
    & \checkmark
    & edge
    & \checkmark
    & \checkmark
    & \checkmark
    & \\
    \cline{2-11}
    &  \multicolumn{1}{|l|}{\shortstack[l]{MLPerf Mobile\tnote{c}}} 
    &
    &
    &
    &
    \checkmark
    &
    mobile 
    & \checkmark 
    &  
    & \checkmark  
    & \\
    \cline{2-11}
    & \multicolumn{1}{|l|}{\shortstack[l]{DeepBench\tnote{d} }}
    &    
    &  
    &  
    & \checkmark 
    & server/edge 
    & \checkmark 
    &  
    &  
    & \\ 
    \cline{2-11}
    &  \multicolumn{1}{|l|}{\shortstack[l]{AI Benchmark\tnote{e}}} 
    &  
    &  
    &
    & \checkmark 
    & mobile 
    & \checkmark 
    &  
    & 
    & \\
    \cline{2-11}
    & \multicolumn{1}{|l|}{\shortstack[l]{EEMBC MLMark\tnote{f} 
    }}
    & 
    & 
    &  
    & \checkmark 
    & edge 
    & \checkmark 
    &  
    & \checkmark 
    & \\ 
    \cline{2-11}
    & \multicolumn{1}{|l|}{\shortstack[l]{AIBench\tnote{g}  }} 
    & \checkmark 
    &  $\triangle$ 
    & \checkmark  
    & \checkmark  
    & server 
    & \checkmark  
    & \checkmark  
    & \checkmark  
    & \checkmark  \\ 
    \cline{2-11}
    & \multicolumn{1}{|l|}{\shortstack[l]{AIoTBench\tnote{h}  }} 
    &  
    &
    &
    & \checkmark 
    & mobile/edge 
    & \checkmark
    & 
    & \checkmark 
    &   \\ 
    \hline

\multicolumn{1}{|c}{\multirow{2}{*}{\textbf{XR}}} 
    & \multicolumn{1}{|l|}{\shortstack[l]{ILLIXR\tnote{i} }}
    & \checkmark  
    &  $\triangle$
    & \checkmark 
    &
    & edge 
    & \checkmark
    & \checkmark
    &  $\triangle$
    & \checkmark \\ 
    \cline{2-11} 
    & \multicolumn{1}{|l|}{VRMark\tnote{j} }
    &     
    &  
    & \checkmark 
    & 
    & PC
    &  
    & 
    & 
    & \checkmark \\ 
    \Xhline{2\arrayrulewidth}
    %
 \multicolumn{1}{|c}{\textbf{ML + XR}} 
    & \multicolumn{1}{|l|}{\textbf{\bench}} 
    & \checkmark 
    & \checkmark 
    & \checkmark
    & \checkmark 
    & edge
    & \checkmark 
    & \checkmark
    & \checkmark 
    & \checkmark\\
\Xhline{2\arrayrulewidth}

 \end{tabular}
      \begin{tablenotes}[flushright , para]\scriptsize
      References:
       \item [a] \cite{reddi2020mlperf},
       \item [b] \cite{banbury2021mlperf},
       \item [c] \cite{reddi2022mlperfmobile},
       \item [d] \cite{Deepbench},
       \item [e] \cite{ignatov2018ai},
       \item [f] \cite{EEMBC_MLMark},
       \item [g] \cite{Gao2019AIBenchAI},
       \item [h] \cite{Luo2018AIoTBT},
       \item [i] \cite{huzaifaillixr},
       \item [j] \cite{VRMark}
       
     \end{tablenotes}
   \end{threeparttable}
  \vspace{-0mm}
\end{table*}

\vspace{-2mm}

\subsection{Why It is Important to Dive into Usage Scenarios}
\label{subsec:eval_scenario}

\vspace{-1mm}


Even though all usage scenarios in \bench reflect the metaverse domain, the individual workload characteristics are diverse and tend to vary during execution, resulting in different system performance. Each usage scenario prefers different accelerator types, as shown in~\autoref{fig:EvalResultsScore4K8K}. For example, in the 4K PE config, the Social Interaction A scenario (\autoref{fig:EvalResultsScore4K8K}, (a)) prefers the FDA style accelerator with WS dataflow (accelerator A). However, Outdoor Activity A (\autoref{fig:EvalResultsScore4K8K}, (c)) prefers the SFDA style with four sub-accelerators with the OS dataflow (accelerator H). 

Moreover, dynamically cascaded models (Section~\ref{sec:challenges}) require a deep dive into corresponding usage scenarios. 
To understand the impact of dynamically cascaded models, we vary the probability of triggering the GE model after the ES (assuming that GE is triggered only if ES results have sufficiently large segmented eyes). \revision{We run 200 experiments and plot the average data to capture the overall trend, focusing on} low- and high-score cases (accelerators B and J) in~\autoref{fig:DynamicCascading}. 

\revision{Overall, both designs maintain their overall scores while we observe a slight decline (0.03) in the overall score on the high-score design (accelerator J) as moving from 25\% to 100\% cascading probability.
As the cascading probability increases from 25\% to 100\%, the real-time score of accelerator B increased by 0.23 points while the QoE score decreased by 0.06 points. This indicates that the low-score design (accelerator B) can drop some frames to maintain overall user experience quantified by \benchscore under high cascading probability. Such results motivate further investigation of optimization opportunities in the scheduler and runtime for XR ML systems.}

\vspace{-2mm}

\subsection{What \revision{the Implications to Future XR Systems} Are}
\label{subsec:eval_hw}

\vspace{-1mm}


\revision{We list three observations we make from the evaluation.}


\noindent
\textbf{\revision{Observation 1) XR systems need to be co-designed with usage scenarios.}} \revision{Evaluation results show that every usage scenario prefers different XR systems. For example, } comparing accelerator styles in the 4K PE setting, we find the accelerator styles with the highest score are all different for each workload. For example, accelerator A (FDA style, WS dataflow, single-accelerator system) is the best style for the social interaction A scenario (\autoref{fig:EvalResultsScore4K8K}, (a)). However, accelerator F (SFDA style, OS dataflow, four-accelerator system) performed the best for the outdoor activity B scenario (\autoref{fig:EvalResultsScore4K8K}, (d)). \revision{Such results suggest that the XR systems require careful co-design with the usage scenarios.}

\noindent
\textbf{\revision{Observation 2) Optimal accelerator styles depend on the chip size.}}
The style H (SFDA style, RS dataflow, four-accelerator system) performs the best for the AR assistant scenario (\autoref{fig:EvalResultsScore4K8K}, (e)) with 4K PEs. However, when the total number of PEs changes to 8K, the style M (HDA style, WS and OS dataflows, four-accelerator system) performs the best. Those results imply that the design space for XR applications is complex with distinctive features of real-time \abbrvWorkload{} workloads, which motivates follow-up studies using \bench.

\noindent
\textbf{\revision{Observation 3) Multi-accelerator systems are friendly to XR workloads.} }
We also find the preference of the number of models in \abbrvWorkload{} models to the multi-accelerator system (e.g., SFDA and HDA). AR assistant (\autoref{fig:EvalResultsScore4K8K}, (e)) and VR gaming (\autoref{fig:EvalResultsScore4K8K}, (f)) scenarios include the most (6) and least (3) number of models, respectively. For AR assistant, we observe the multi-accelerator style (SFDA and HDA) outperforms the single accelerator style. For VR gaming scenario, in contrast, the FDA style (accelerator A) outperforms most of the other accelerators. In particular, when the sub-accelerator size is sufficiently large (8K PE), a quad-accelerator system (HDA accelerator M) performs the best on the many-model scenario (AR assistant), but the same system underperforms on the fewer-model scenario (VR gaming). Such data show the efficacy of parallel model execution using
sub-accelerators, which motivates to explore scale-out designs for many-model \abbrvWorkload{} workloads like the AR assistant.

\vspace{-2mm}

\section{Related Work}
\label{sec:related_work}

\vspace{-1mm}

Based on the characteristics we describe in \autoref{sec:benchmark}, we present the limitations of existing ML and XR benchmarks in \autoref{tab:related_works}. \bench is unique in that it is the only workload suite that captures complex workload dependencies, is ML-focused, presents several real-world usage scenarios that are distilled from industry practice and uniquely establishes a robust scoring metric. Due to space limitations, we defer detailed discussions of the benchmarks to \autoref{sec:related_work_appendix}. In summary, \bench is the first suite to include several ML workloads tailored for XR applications.

\vspace{-2mm}

\section{Conclusion}
\label{sec:future_works}

\vspace{-1mm}

Metaverse use cases necessitate complex ML benchmark workloads that are essential for fair and useful analyses of existing and future system performance, but such workloads exceed the capabilities of existing benchmark suites. The XR benchmark we present, which is based on industry experience, captures the diverse and complex characteristics of these emerging ML-based \abbrvWorkload{} workloads. We believe \bench will foster new ML systems research focused on XR.

\section*{Acknowledgements}

This work was enabled in part by support from Robert Shearer at Meta. We appreciate Rob's ongoing assistance and counsel in helping us create methodical approaches to assess XR SoC designs. Part of the funding for authors from Harvard University came from the Office of the Director of National Intelligence (ODNI), Intelligence Advanced Research Projects Activity (IARPA) and the Semiconductor Research Corporation (SRC). 
The Georgia Tech authors were funded in part by SRC.
The views and conclusions contained herein are those of the authors and should not be interpreted as necessarily representing the official policies, either expressed or implied, of ODNI, IARPA, SRC, or the U.S. Government. The U.S. Government is authorized to reproduce and distribute reprints for governmental purposes notwithstanding any copyright annotation therein. 


\bibliography{ref}
\bibliographystyle{mlsys2023}

\clearpage
\appendix
\section{Benchmark Model Instances}
\label{sec:model_instances}

\begin{table*}[t]
\centering
\setlength{\abovecaptionskip}{0pt}
\setlength{\belowcaptionskip}{0pt}
\scriptsize
\caption{\small Specific model instances for the \bench unit models listed in \autoref{tab:unit_models}. Also classifies model type and major operators.}
\label{tab:model_instances_ops}
\begin{tabular} {|@{~} c @{~}|l|l|l|l|l|}
\hline
\multicolumn{1}{|c|}{\textbf{Task}} &
\multicolumn{1}{c|}{\textbf{Model Reference}} &
\multicolumn{1}{c|}{\textbf{Model Instance}} &
\multicolumn{1}{c|}{\textbf{Model Type}} &
\multicolumn{1}{c|}{\textbf{Major Operators}}
\\
\hline

\hline

HT
& Hand Graph-CNN~\cite{ge2019handshapepose}
& Hand Shape/Pose~\cite{ge2019handshapepose}
& CNN
& CONV2D, Maxpool, FC
\\
\hline
ES
& RITNet~\cite{chaudhary2019ritnet} 
& RITNet~\cite{RITNetCode}
& CNN
& CONV2D, AvgPool, Skip Connection
\\
\hline
GE
& Eyecod~\cite{you2022eyecod}
& FBNet-C~\cite{FBNetCCode} 
& CNN
& CONV2D, DWCONV, Skip Connection
\\
\hline
KD
& Key-Res-15~\cite{tang2018deep}
& res8-narrow~\cite{tang2018deep}
& CNN
& CONV2D, Avgpool, Skip Connection
\\
\hline
SR
& Emformer ~\cite{shi2021emformer}
& EM-24L~\cite{shi2021emformer}
& Transformer
& Self-attention, Layernorm
\\
\hline
SS
& HRViT~\cite{gu2022multi}
& HRViT-b1~\cite{HRViT-b1}
& Transformer
& Self-attention, Layernorm, DWCONV
\\
\hline
OD
& D2go~\cite{Faster-RCNN-FBNetV3A}
& Faster-RCNN-FBNetV3A~\cite{Faster-RCNN-FBNetV3A}
& R-CNN
& CONV2D, DWCONV, Skip Connection
\\
\hline
AS
& TCN ~\cite{lea2017temporal}
& ED-TCN~\cite{ED-TCN}
& CNN
& CONV2D, Maxpool, Upsample
\\
\hline
DE
& MiDaS~\cite{ranftl2020towards}
& midas\_v21\_small~\cite{midasV21Small}
& CNN
& CONV2D, DWCONV, Skip Connection
\\
\hline
DR
& Sparse-to-Dense~\cite{ma2018sparse}
& RGBd-200~\cite{ma2018sparse}
& CNN
& CONV2D, DeCONV, DWCONV
\\
\hline
PD
& PlaneRCNN~\cite{liu2019planercnn}
& PlaneRCNN~\cite{liu2019planercnn}
& R-CNN
& CONV2D w/ FPN, RoIAlign
\\ 
\hline
 \end{tabular}
\end{table*}


As an extension to \autoref{tab:unit_models}, this section describes more details on the models included in \bench. \autoref{tab:unit_models} specifies which model variation (Model Instance) is adopted from the representative model (Model Reference), along with baseline or backbone structure (Model Type), and types of major operators that compose the model (Major Operators). The model instances are chosen based on their size, considering the edge use case. 
In addition, we also down-scale the dataset resolution of certain tasks to adjust to the context of edge devices. Stereo Hand Pose \cite{stereohandpose} is scaled by $1/2$ for Hand Tracking (HT), OpenEDS 2019 \cite{garbin2019openeds} and OpenEDS 2020 \cite{palmero2021openeds2020} are both scaled by $1/4$ for Eye Segmentation (ES) and Gaze Estimation (GE), respectively, and KITTI \cite{geiger2012we} is scaled by $1/4$ for Plane Detection (PD).

As shown in the table, there is a variety of model types and operators included in the \bench workloads, representative of the diverse computing requirements of an XR system. Such heterogeneity emphasizes the need for innovative solutions to realize XR device capabilities.



\section{Problem Formulation}
\label{sec:appendix_formulation}


We formulate the benchmark and scores using symbols presented in~\autoref{tab:symbol_table} and~\autoref{box:base_defs}. We provide more details about the definitions in~\autoref{box:base_defs} and~\autoref{box:score_metrics} in this section.

\begin{definition}
\textbf{Input Data Stream ($St_{input}$)} \\
The input data stream $I$ is defined as follows:
$$St_{input} = \{\sigma ~|~ \sigma = (inSrc_{ID}, FPS_{sensor}, L_{init}, Jt)\}$$
\label{def:input_data_stream}
\vspace{-5mm}
\end{definition}
~\autoref{def:input_data_stream} formulates the input stream description in~\autoref{tab:input_sources}. The $inSrc_{ID}$ refers to a string value that refers to the input source identifier. $FPS_{sensor}$ refers to the streaming rate (FPS) of the associated sensor. $L_{init}$ refers to the initial latency of each input stream, and $Jt$ refers to the maximum jitter in milliseconds.

\begin{definition}
\textbf{Model Quality Goal (Q)} \\
$$Q = (QM_{ID}, QM_{Targ}, QM_{Type})$$
\label{def:model_quality_goal}
\vspace{-5mm}
\end{definition}

Model quality refers to the degree of achieved target metrics (e.g., mIoU and accuracy) of each model. $QM_{ID}$ refers to the name of the metric, $QM_{Targ}$ refers to the float point value representing the target value of the metric (e.g., 0.96 for classification accuracy), and $QM_{Type}$ indicates if the metric is higher or lower-is-better (HiB or LiB) metric.

\begin{definition}
\textbf{Unit models ($M$)} \\
The set of unit models $M$ is defined as follows:
$$ M = \{\mu ~|~ \mu \in (M_{ID}, DS_{ID}, \sigma, Q) \wedge \sigma \in St_{input} \}$$
\label{def:unit_models}
\end{definition}

~\autoref{def:unit_models} defines a set of unit models utilized in \bench to construct complex usage scenarios. $\mu$ refers to a unit model (i.e., an element of $M$), $M_{ID}$ refers to the model name in string, $DS_{ID}$ refers to the name of associated dataset, $\sigma$ refers to an input stream from a sensor associated with the unit model, $Q$ is the model quality goal defined in~\autoref{def:model_quality_goal}. If a model utilizes multi-modality inputs, $\sigma$ becomes a set of associated input streams. Based on the definition of $M$, we define the usage scenario ($\theta$) as follows:

\begin{definition}
\textbf{Usage Scenario ($\theta$)} \\
$$
\theta = \{(\mu, Dep_{\mu}, FPS_{model})~|~\mu \in M \wedge Dep_{\mu} \subset M  \} 
$$
\label{def:usage_scenario}
\vspace{-6mm}
\end{definition}

In \autoref{def:usage_scenario}, $Dep_{\mu}$ defines the model granularity dependency on $\mu$, which is a list of models on which $\mu$ depends. With~\autoref{def:usage_scenario}, we can define the benchmark suite as follows:

\begin{definition}
\textbf{Benchmark Suite ($\Omega$)} \\
Given a set of usage scenarios $\Theta$, a real-time MTMM benchmark suite  $\Omega$ is defined as follows: 
$$
\Omega = \{\theta_{1}, \theta_{2}, ... \theta_{NumScn}\} 
$$
\label{def:becnmark_suite}
\vspace{-6mm}
\end{definition}

The~\autoref{def:becnmark_suite} shows that a real-time MTMM benchmark is a collection of usage scenarios as described in~\autoref{tab:usage_scenario_processing_rates}. $NumScn$ refers to the number of usage scenarios \bench includes.

Based on the formulation on workload side so far, we define some additional concepts for defining \bench's scoring metrics.

\begin{definition}
\textbf{Inference Request ($IR$)} \\
$$
IR = (\mu, InFrame_{ID})
$$
\label{def:inference_request}
\vspace{-5mm}
\end{definition}

Using~\autoref{def:unit_models} and~\autoref{def:inference_request}, we define the inference request time and deadline as follows:

\begin{definition}
\textbf{Inference Request time ($T_{req}(IR)$)} \\
$$
T_{req}(IR) = L_{init}(inSrc_{ID}) + \frac{InFrame_{ID}}{FPS_{Sensor}(inSrc_{ID})}$$ 
$$+ 2 Jt ~(Dist(rand(inSrc_{ID}\times InFrame_{ID})) - 0.5) $$
$$
where ~Dist(x) \in [0,1]~ \wedge ~x \in \mathbb{R}
$$
\label{def:inference_request_time}
\end{definition}

$L_{init}(inSrc_{ID})$ indicates the setup latency of the input stream from the input source $inSrc_{ID}$. $InFrame_{ID} \times 1/FPS_{Sensor}(inSrc_{ID})$ represents the time until an XR device reaches the $InFrame_{ID}$ frame under the streaming rate (FPS) of a corresponding input stream $\sigma$.
The term, $Jt \times (Dist(rand(inSrc_{ID} \times InFrame_{ID}) - 0.5$, accounts for the impact of jitters on the arrival time modeled by a maximum jitter $Jt$, a distribution $Dist(x) \in [0,1]$ for $x \in \mathbb{R}$, and a random function $rand = f : \mathbb{N} \rightarrow \mathbb{R}$. Note that we make the choice of $Dist(x)$ and $rand(n)$ flexible for various scenarios. By default, $Dist(x)$ is a Gaussian distribution and $rand(n)$ is the rand function of C++17 standard library, \textit{cstlib}.

Using a similar formulation, we define the inference deadline as follows.

\begin{definition}
\textbf{Inference Deadline ($T_{dl}(IR)$)} \\
The deadline for an inference request $IR$ is defined as follows:
$$
T_{dl}(IR)= L_{init}(inSrc_{ID}) + \frac{InFrame_{ID} + 1}{SR(inSrc_{ID})}
$$
\label{def:deadline}
\end{definition}

The definition in~\autoref{def:deadline} indicates that the deadline of an inference on an input frame is the arrival time of the next input frame.

\begin{definition}
\textbf{Inference Slack ($T_{sl}(IR)$)} \\
The inference slack, the length of time window given for an inference run associated with $IR$ is defined as follows:
\begin{align*}   
T_{sl}(IR)= T_{dl}(IR) - T_{req}(IR)
\end{align*}
\label{def:slack}
\end{definition}

Using the definitions from ~\autoref{def:input_data_stream} to ~\autoref{def:slack}, we define \bench's score metrics.

\subsection{Inference-level Benchmark Score}
\label{sec:inference_wise_metric}

Because of the real-time processing nature of the metaverse workloads, it is challenging to compare metaverse device systems running real-time MTMM using traditional metrics of using latency and energy. The latency measures the end-to-end execution time of each inference, which can be used to check if each model's deadlines are satisfied. However, achieving less latency than the deadlines does not offer benefits, making latency not an absolute minimization goal, unlike other ML systems targeting non-real-time MTMM workloads. In addition, we can adjust energy to meet the deadlines or optimize using the slack to the deadline (e.g., DVFS), which also makes energy a knob, not an absolute minimization target.

Therefore, we explore a set of new metrics for \bench that considers all the unique aspects of real-time MTMM workloads we discussed: (1) the task-specific deadlines based on usage scenarios, (2) end-to-end latency (i.e., how much latency does an ML system need beyond the deadline?), (3) overall energy consumption, and (4) the quality of experience delivered. 

To facilitate an intuitive comparison of ML systems with many pillars, we propose a single score metric that captures all of the above. The single-score approach will also help motivate the industry to submit their results because the latency and energy can be confidential data that cannot be directly shared with the public. Instead, the industry can offer the single score metric capturing overall performance on real-time and multi-model DL workloads to demonstrate the robust capabilities of their accelerator systems.


To construct such a metric, we consider each aspect of real-time MTMM workloads and model performance (e.g., accuracy, mIoU, and boxAP): real-time requirement, energy consumption, quality of experience, and model performance. We define a score for each pillar and combine them to define the single comprehensive metric. We define those score functions to be in [0, 1] range to facilitate component-wise analysis as well.

To model real-time requirements, we consider the following observations: (1) Too much optimization on latency beyond the deadline does not lead to higher processing rates; even if a system finished the latency in only one cycle, the system still needs to wait for the next input frame. (2) reduced latency can still be helpful for scheduling other models. (3) violated deadline gradually damages the user experience (e.g., Achieving 59Hz for an eye-tracking model targeting 60Hz wouldn't significantly affect the user experience).

Based on those observations, we search for a function that (1) gradually rewards and penalizes the reduced and increased latency near a deadline (e.g., $\pm$ 0.5ms for a deadline of 10ms) and (2) outputs 0 and 1 if the latency is beyond or within the deadline. We discuss such a function in~\autoref{def:realtime_score_appendix}.

\begin{definition}
\textbf{Real-time (RT) score} \\
For an inference request IR, the real-time score is defined as follows:
$$
RtScore(IR) = \frac{1}{1+e^{k(L_{Inf}(IR)-T_{sl}(IR))}}
$$
\label{def:realtime_score_appendix}
\end{definition}

The definition of $RtScore(IR)$ in~\autoref{def:realtime_score_appendix} is based on the inference latency $L_{Inf}(IR)$ and the time window given for the inference $IR$, $T_{sl}(IR)$.

The RT score function can also be viewed as a modified Sigmoid function with shifting and scaling. A benefit of using $RtScore$ is that we can tune the constant $k$ depending on the deadline sensitivity. ~\autoref{fig:RealtimeScoreFunc} shows the change of the $RtScore$ based on the k value, and we can observe large $k$ values makes the function more sensitive around the deadline. We set 15 as the default value of $k$ and utilize it in our evaluation.  

\insertFigure{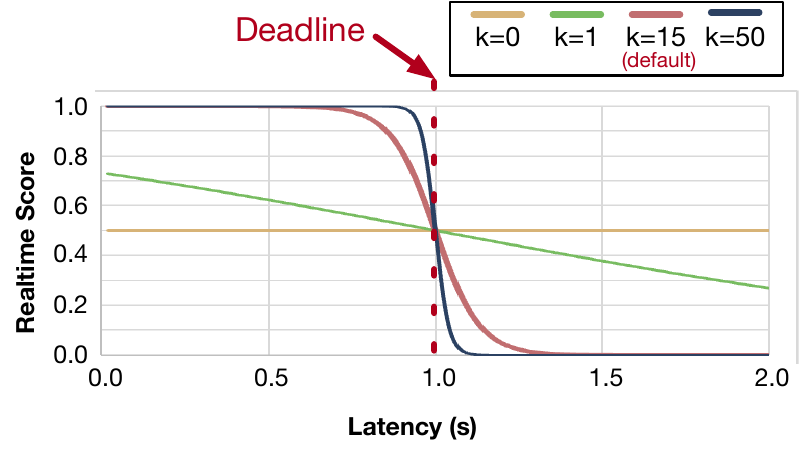}{An example real-time score function over different values of the parameter $k$ whose range is $[0,\infty)$. We assume the time window between the inference request time and deadline to be 1 s in this example for simplicity.  If k is 0, the score is completely not relevant to the deadline (i.e., no sensitivity on deadline). If k is $\infty$, the score function becomes a piece-wise function that flips the score from 1 to 0 at the deadline.}


\begin{definition}
\textbf{Energy (En) score} \\
For an inference request IR, the energy score is defined as follows:
$$
EnScore(IR) = \frac{En_{max}-En(IR)}{En_{max}}
$$
\label{def:energy_score_appendix}
\end{definition}

To make the energy score as higher-is-better metric, consistent with other scores, we utilize $En_{max}$, which represents the maximum energy allowed for each inference. By default, we set $En_{max}$ as 1500 mJ.
\\
\\

\begin{definition}
\textbf{Accuracy (Acc) score} \\
\begin{align*}
AccScore(IR) =& max(1, rawAccScore(IR)) \\ 
rawAccScore(IR) =& 
    \begin{cases}
      \frac{QM_{measured}}{QM_{targ}},&\text{if}\ QM_{Type} = HiB\\
      \frac{QM_{targ}}{QM_{measured}+\epsilon},& \text{otherwise}
    \end{cases}\\
    & \text{where } \epsilon > 0 \wedge \epsilon \ll 1 \wedge \epsilon \in \mathbb{R}
\end{align*}
\label{def:accuracy_score_appendix}
\end{definition}

For an inference request $IR$, the accuracy (Acc) score quantifies the model quality (or model performance) as a value within the [0,1] range. Depending on the model quality type (higher- and lower-is-better), Acc score is computed in a different way to formulate the Acc score as a higher-is-better metric. For lower-is-better model quality metric, we utilize a small real number $\epsilon$ for numerical stability, preventing divide by zero errors. By default, $\epsilon$ is set as $10^{-6}$. 

The three base scores, RT, EN, and Acc Scores, are defined for each inference runs. Unlike those, the next score, quality of experience score is defined for entire inference runs for a model.

\begin{definition}
\textbf{Quality of Experience (QoE) score} \\
For a unit model $\mu$, the quality of experience (QoE) score is defined as follows:
$$
QoEScore(\mu) = \frac{NumFrm_{exec}(\mu)}{NumFrm(\mu)}
$$
\label{def:qoe_score_appendix}
\end{definition}

$NumFrm(\mu)$ refers to the total number of input frames for a unit model $\mu$ streamed during the entire execution of the workload. $NumFrm_{exec}(\mu)$ refers to the number of actually processed input frames using $\mu$. 

The QoE score quantifies the ratio of the number of  processed input frames and the total number of streamed input frames for a model $\mu$. The QoE score is defined in a usage scenario granularity because frame drop can be measured for the entire inference runs for a model.

Using four unit scores (RT, En, Acc, and QoE scores,), we formulate the inference, usage scenario, and benchmark scores.

\begin{definition}
\textbf{Inference-wise score}
\begin{align*}    
Score_{inf}(IR) = &RtScore(IR) \times EnScore(IR) \\  
&\times AccScore(IR)
\end{align*}
\label{def:inference_score}
\end{definition}

As illustrated in~\autoref{fig:ScoresOverview}, we compute the product of three unit scores defined in the inference granularity to define an aggregated metric for an inference. Combining the inference-wise score with the QoE score, we construct the usage scenario granularity score.

\begin{definition}
\textbf{Usage-scenario Score}
\begin{align*}
Score_{scn}(\theta) =
\sum_{j=1}^{NumFrm(\mu)}{} 
\frac{Score_{inf}(IR)\times QoEScore(\mu)}{NumFrm(\mu) \times |\theta|}    
\end{align*}
\label{def:usage_scenario_score_function_appendix}
\end{definition}

Using usage-scenario-wise scores, we define an overall score for \bench, \benchscore, as the average of the scores for each usage scenario in \bench. 

\begin{definition}
\textbf{\benchscore}
\begin{align*}
Score_{bench} = \frac{\sum_{\theta \in \Omega}{Score_{scn}(\theta)}}{|\Omega|}
\end{align*}
\label{def:xrbench_score_function_appendix}
\end{definition}

As shown in~\autoref{def:xrbench_score_function_appendix}, \benchscore ($Score_{bench}$) summarizes scores for each usage scenario using average.

\subsection{Schedule}

We do not propose a specific scheduler as it is a part of the ML system software to be evaluated. However we define valid schedules to satisfy the following conditions:

\insertFormula{Dependency Condition}
{-1.5}
{
\forall h_1, h_2 \in HW, \mu_{i} \in Dep_{\mu_{j}}, T_{end}(\mu_{i}, h_1) < T_{start}(\mu_{j}, h_2)
}
\insertFormula{Hardware Occupancy Condition}
{-3}
{
&\forall \mu_{1}, \mu_{2} \in M, \\ 
&(T_{end}(\mu_{1},h) \leq T_{start}(\mu_{2},h)) \lor (T_{end}(\mu_{2},h) \leq T_{start}(\mu_{1},h)) \\ 
&\text{where } T_{end}(h,\mu) = \infty \land T_{start}(h,\mu) = 0 \text{ if } \mu \text{ is not mapped on } h
}

The dependency condition indicates that the dependency order must be maintained in a schedule. The hardware occupancy condition indicates that a hardware piece (e.g., a systolic array-based accelerator) cannot run two models simultaneously. That is, if a hardware piece can run multiple models simultaneously, it should be treated as multiple smaller hardware pieces. 

\section{Detailed Related Work Comparison}
\label{sec:related_work_appendix}

In this section, we expand on \autoref{sec:related_work} and \autoref{tab:related_works} by providing detailed discussions on prior benchmarks. 

\betterparagraph{General ML Workload Benchmarks}
 MLPerf Inference ~\cite{reddi2020mlperf} is a set of industry standard, single-kernel ML benchmarks that span the ML landscape, from high performance computers~\cite{farrell2021mlperf} to tiny embedded systems~\cite{banbury2021mlperf}. It also provides a rich set of inference scenarios based on realistic use cases from industry: single-stream (single inference), multistream (repeated inference with a time interval), server (random inference request modeled via Poisson distribution), and offline (batch processing). Extensions to embedded systems (MLPerf Tiny ~\cite{banbury2021mlperf}) and mobile devices like smartphones (MLPerf Mobile ~\cite{reddi2022mlperfmobile}) have also been developed, drawing closer to the XR form factor. However, the MLPerf suite workloads do not deploy models in a concurrent or cascaded manner and the scoring metrics lack QoE consideration, which are essential in XR workloads. 

DeepBench~\cite{Deepbench} focuses on benchmarking kernel operations which underlie ML performance. Although such microbenchmarks provide insights to operator level optimizations, it cannot be used for understanding the end-to-end performance of a single model or for \abbrvWorkload {} workloads. AI Benchmark~\cite{ignatov2018ai} targets the ML inference performance of smartphones with 14 different tasks and EEMBC MLMark~\cite{EEMBC_MLMark} measures the performance of neural networks on embedded devices. Still, none of them cover \abbrvWorkload {} performance nor consider real-time processing scenarios. Their scoring metrics are also not sufficiently diverse to handle complex XR workloads. 

AIBench~\cite{Gao2019AIBenchAI} from BenchCouncil is another industry standard AI benchmark for Internet services, which was one of the first to include application scenarios for end-to-end performance evaluation. These scenarios model \abbrvWorkload {} workloads of E-commerce search intelligence use cases with heterogeneous latency of each model, provided with rich scoring metric components for evaluation. Although AIBench decently reflects the key components of real-time \abbrvWorkload {} workloads, the benchmark is tailored to server-scale internet service and has little to do with edge applications. In addition, their static execution graphs make extensions to XR use cases difficult, which require dynamic execution of models based on their control dependencies. 


AIoT~\cite{Luo2018AIoTBT, AIoIBT} is an AIBench extension that focuses on mobile and embedded AI. Though these platforms come closer to the XR platform, the benchmark does not model real-time, \abbrvWorkload-based scenarios and therefore falls short to serve as an XR benchmark.

\betterparagraph{XR Benchmarks}
 ILLIXR~\cite{huzaifaillixr} is a benchmark suite tailored for  XR systems. ILLIXR models concurrent and cascaded execution pipelines in XR use cases and considers the real-time requirements of XR devices. Although ILLIXR provides a solid benchmark in the XR domain, the focus of ILLIXR is mainly in non-ML-based pipelines, unlike the ML workload focus of \bench. ILLXR includes one ML model (RITNet for eye tracking), and its other parts are based on traditional computer vision and audio algorithms (e.g., QR decomposition and Gauss-Newton refinement) and signal processing (e.g., FFT). 

VRMark~\cite{VRMark} is a benchmark that evaluates the performance of VR experiences on PCs. The benchmark also does not target ML performance assessments but rather focuses on rendering graphics. Moreover, it lacks usage scenarios that are reflective of real-world user characteristics and various score metrics for systematical analysis.

\betterparagraph{ML and XR Benchmarks}
Compared to the above-mentioned benchmarks ~\cite{reddi2020mlperf, farrell2021mlperf,banbury2021mlperf,ignatov2018ai,Deepbench, EEMBC_MLMark,Gao2019AIBenchAI, Luo2018AIoTBT, huzaifaillixr, VRMark}, \bench covers all requirement of an ML-based XR workloads. To be specific, \bench provides diverse cascon-\abbrvWorkload {} scenarios with real-time requirement and complex dependencies, which majority ML benchmarks are missing. Careful consideration of QoE aspects in XR applications into its scoring metric is another strength of \bench that distinguishes it from other prior works.  

 On the other hand, even though existing XR-related or scenario-based benchmarks support real-time \abbrvWorkload {} scenario and QoE metrics, they still lack several components such as sufficient ML algorithm coverage, dynamic model execution graph, and focusing on edge devices. All of these characteristics are satisfied by \bench, expecting significant contribution to XR research community and the industry.


\section{Artifact Appendix}
\subsection{Abstract}

This appendix describes the complete workflow for running XRBench-MAESTRO and generating results used in the paper. 

\subsection{Artifact check-list (meta-information)}
{\small
\begin{itemize}
  \item {\textbf{Algorithm:} Scheduling is based on a latency-greedy scheduler; dispatch an inference job to an idle accelerator with the minimal expected latency.}
  \item {\textbf{Program:} Based on MAESTRO (\url{https://maestro.ece.gatech.edu})}
  \item {\textbf{Compilation:} C++ compilers that support C++17 or later (tested compilers: clang-1400.0.29.202 or g++ (Ubuntu 11.3.0-1ubuntu1~22.04))}
  \item {\textbf{Run-time environment:} Tested environments: MacOS 13.2.1 (22D68) and Ubuntu 22.04}
  \item {\textbf{Hardware:} X86-64 processor-based Linux machines, X86-64 processor-based Mac Machines (e.g., Mackbook Pro and iMac), or Apple Silicon-based Mac Machines (e.g., Macbook pro with M1 processor)}
  \item {\textbf{Execution:} Automated scripts are included (please refer to README in the code for details) }
  \item {\textbf{Metrics: } score metrics proposed in the paper}
  \item {\textbf{Output: } Plots in pdf files (under /XRbench\_evaluation/plots), data in csv files (under /XRbench\_evaluation/eval\_data)}
  \item {\textbf{How much disk space required (approximately)?:} 10-20 MB}
  \item {\textbf{How much time is needed to prepare workflow (approximately)?:} expected to be less than 30 minutes. }
  \item {\textbf{How much time is needed to complete experiments (approximately)?:} Depends on the machine's computing power. On our tested machines with latest processors (e.g., Intel i9-13900k with 128GB RAM and Apple M1 Pro with 64 GB RAM), the experiments overall take less than 30 minutes.}
  \item {\textbf{ Publicly available?:} doi: \url{10.5281/zenodo.7857382}, Github: \url{https://github.com/XRBench}}
\end{itemize}

\subsection{Description}

\subsubsection{How delivered}

A Dropbox download link of a zip file of the code will be shared with the AE reviewers (please see ``abstract" in the Hotcrp submission). We will open-source a clean version along with the conference.

\subsubsection{Hardware dependencies}

Typical desktop, laptop, or server running Linux or Mac OS are required. We tested the artifact on X86-64 processor-based Linux machines, X86-64
processor-based Mac Machines (e.g., Mackbook Pro and iMac),
and Apple Silicon-based Mac Machines (e.g., Macbook pro with
M1 processor)

\subsubsection{Software dependencies}
\begin{itemize}
    {\item C++ compiler supporting C++17 (tested compilers: clang and g++)}
    {\item scons}
    {\item boost library (C++) }
    {\item Python3 (3.10 or later)}
    {\item matplotlib}    
\end{itemize}
\subsubsection{Data sets}

\subsection{Installation}

\textbf{SW dependency installation guide for Linux machines (Ubuntu)}
\begin{verbatim}
> sudo apt-get install g++ libboost-all-dev \ 
                       python3 scons python3-pip
> pip3 install matplotlib
\end{verbatim}

\noindent
\textbf{SW dependency installation guide Guide for Mac machines (Using Homebrew)}
\begin{verbatim}
> brew install scons python3 boost
> pip3 install matplotlib
\end{verbatim}

\noindent
Note: Apple silicon-based Mac machines have issues in linking boost library during compilation. An alternative SConstruct file (specification file for scons-based compilation flow) that addresses this issue is included in the codebase; please refer to README file for details.)


\subsection{Experiment workflow}

\begin{verbatim}
    > scons
    > ./reproduce_figure5.sh
    > ./reproduce_figure6.sh
    > ./reproduce_figure7_data.sh    
\end{verbatim}


\subsection{Evaluation and expected result}

\noindent
\textbf{Reproducing Figure 5}
Running ``reproduce\_figure5.sh,"  XRBench-MAESTRO will reproduce plots in Figure 5 of the original paper under ``XRbench\_evaluation/plots/4K" and ``XRbench\_evaluation/plots/8K."
Please note that results on outdoor activity A, outdoor activity B, and AR assistant are non-deterministic (the workload is dynamic). Due to such dynamic workloads, ``gross\_average" plots will look slightly different as well. You can still expect exact match of results on social interaction A, social interaction B, and AR gaming.

\textbf{Reproducing Figure 6}
Running ``reproduce\_figure6.sh" after ``reproduce\_figure5.sh," XRBench-MAESTRO will reproduce plots in Figure 6 of the original paper under ``XRbench\_evaluation/plots." You can expect exact match of the results. Please note that the aspect ratio of the figure 6 and generated plots are different (and Figure 6 has mis-aligned x-axis tick labels), but they will show the same data.

\textbf{Generating Figure 7 data}
Running ``reproduce\_figure7\_data.sh" after ``reproduce\_figure5.sh," and ``reproduce\_figure6.sh," data under Figure 7 will be generated under ``XRbench\_evaluation/eval\_data/figure7/."

Please expect some fluctuations in results as the workload for Figure 7 is dynamic.

\subsection{Experiment customization}

The provided automated scripts cover entire evaluation we ran. Users can change settings by modifying the contents of files under the followings.

\begin{itemize}
    {\item \textbf{Modifying dataflow styles of accelerators: } Please modify ``XRbench\_evaluation/dataflows" to change processing style of accelerators (i.e, dataflow). The description is based on MAESTRO (https://maestro.ece.gatech.edu) style dataflow notation.}
    {\item \textbf{Modifying hardware styles: } Please modify ``XRbench\_evaluation/hw\_configs" to change hardware parameters (e.g., number of PEs, number of sub-accelerators in the hardware system, etc.) }
\end{itemize}



\end{document}